\documentclass[lettersize,journal]{IEEEtran}
\usepackage{amsmath,amsfonts}
\usepackage{algorithmic}
\usepackage{algorithm}
\usepackage{array}
\usepackage[caption=false,font=normalsize,labelfont=sf,textfont=sf]{subfig}
\usepackage{textcomp}
\usepackage{stfloats}
\usepackage{url}
\usepackage{verbatim}
\usepackage{graphicx}
\usepackage{cite}
\usepackage{amsmath}
\usepackage{booktabs}
\usepackage{bm}
\usepackage{amssymb}
\usepackage{multirow}
\usepackage{siunitx}
\usepackage{threeparttable}

\begin{document}

\title{Human-Level Control through Directly-Trained Deep Spiking Q-Networks}

\author{Guisong~Liu,
        Wenjie~Deng,
        Xiurui~Xie,
        Li~Huang,
        Huajin~Tang
\thanks{Guisong Liu, Li Huang are with the School of Computing and Artificial Intelligence, South-western University of Finance and Economics, Chengdu, 611130, China, and Guisong Liu is also with School of Computer Science, Zhongshan Institute, University of Electronic Science and Technology of China, Zhongshan, 528400, China (email: gliu@swufe.edu.cn).}
\thanks{Wenjie Deng, Xiurui Xie are with the School of Computer Science and Engineering, University of Electronic Science and Technology of China, Chengdu, 611731, China. (\textit{Corresponding author}: Xiurui Xie, email: xiexiurui@uestc.edu.cn)}
\thanks{Huajin Tang is with the College of Computer Science and Technology, Zhejiang University, Hangzhou 310027, China, and also with Zhejiang Laboratory, Hangzhou 311122, CHINA (email: htang@zju.edu.cn).}
\thanks{This work was supported by the Natural Science Foundation of Guangdong Province under Grant No. 2021A1515011866 and Sichuan Province under Grant No. 2021YFG0018 and No. 2022YFG0314, and the Social Foundation of Zhongshan Sci-Tech Institute under Grant No. 420S36.}
\thanks{Manuscript received XX, X; revised XX, X.}}

\markboth{Journal of X,~Vol.~X, No.~X, X~X}%
{Shell \MakeLowercase{\textit{et al.}}: Human-Level Control through Directly-Trained Deep Spiking Q-Networks}


\maketitle

\begin{abstract}
As the third-generation neural networks, Spiking Neural Networks (SNNs) have great potential on neuromorphic hardware because of their high energy-efficiency.
However, Deep Spiking Reinforcement Learning (DSRL), i.e., the Reinforcement Learning (RL) based on SNNs, is still in its preliminary stage due to the binary output and the non-differentiable property of the spiking function.
To address these issues, we propose a Deep Spiking Q-Network (DSQN) in this paper.
Specifically, we propose a directly-trained deep spiking reinforcement learning architecture based on the Leaky Integrate-and-Fire (LIF) neurons and Deep Q-Network (DQN).
Then, we adapt a direct spiking learning algorithm for the Deep Spiking Q-Network.
We further demonstrate the advantages of using LIF neurons in DSQN theoretically.
Comprehensive experiments have been conducted on 17 top-performing Atari games to compare our method with the state-of-the-art conversion method.
The experimental results demonstrate the superiority of our method in terms of performance, stability, generalization and energy-efficiency.
To the best of our knowledge, our work is the first one to achieve state-of-the-art performance on multiple Atari games with the directly-trained SNN.
\end{abstract}


\section{Introduction}
\IEEEPARstart{I}{n} recent years, Spiking Neural Networks (SNNs) have attracted widespread interest because of their low power consumption \cite{Ju} on neuromorphic hardware.
In contrast to Artificial Neural Networks (ANNs) that use continuous values to represent information, SNNs use discrete spikes to represent information, which is inspired by the behavior of biological neurons in both spatiotemporal dynamics and communication methods.
This makes SNNs has been implemented successfully on dedicated neuromorphic hardware, such as SpiNNaker at Manchester, UK \cite{Furber}, IBM's TrueNorth \cite{Merolla}, and Intel's Loihi \cite{Davies}, which are reported to be 1000 times more energy-efficient than conventional chips.
Furthermore, recent studies demonstrated competitive performance of SNNs compared with ANNs on image classification \cite{Wu}, object recognition \cite{Wang, Kim}, speech recognition \cite{Morales, Wu3}, and other fields \cite{Guo, Yu, Aquino, Gerardo, Liu, Wu1}.

The present work focuses on combining SNNs with Deep Reinforcement Learning (DRL), i.e., Deep Spiking Reinforcement Learning (DSRL), on Atari games.
Compared to image classification, DSRL on Atari games involve additional complexity due to the pixel image as input and the partial observability of the environment.
The development of DSRL lags behind DRL, while DRL has made tremendous successes, achieved and even surpassed human-level performance in many Reinforcement Learning (RL) tasks \cite{Mnih1, Mnih, Silver, Silver1, Vinyals, Hessel}.
The main reason is that, training SNNs is a challenge, as the event-driven spiking activities are discrete and non-differentiable.
In addition, the activities of spiking neurons are propagated not only in spatial domain layer-by-layer, but also along temporal domain \cite{Pfeiffer}.
It makes the training of SNNs in reinforcement learning more difficult.

To avoid the difficulty of training SNNs, \cite{Carrasco} proposed an alternative approach of converting ANNs to SNNs.
\cite{Devdhar} extended existing conversion methods \cite{Diehl, Rueckauer, Sengupta} to the domain of deep Q-learning, and improved the robustness of SNNs in input image occlusion.
After that, \cite{Tan} proposed a more robust and effective conversion method which converts a pre-trained Deep Q-Network (DQN) to SNN, and achieved state-of-the-art performance on multiple Atari games.
Nevertheless, the existing conversion methods rely on pre-trained ANNs heavily.
Besides, they require very long simulation time window (at least hundreds of timesteps) for convergence, which is demanding in terms of computation.

To maintain the energy-efficiency advantage of SNNs, the direct training methods have been widely studied recently \cite{Xie, Xiurui, Fang, Xu, Wu2, Zhang}.
For instance, \cite{Zheng} proposed a threshold-dependent batch normalization method to train deep SNNs directly.
It firstly explored the directly-trained deep SNNs with high performance on ImageNet.
Besides, Surrogate Gradient Learning (SGL) is proposed to address the non-differentiable issue in spiking function by designing a surrogate gradient function that approximates the spiking back-propagation behavior \cite{Neftci}.
The flexibility and efficiency make it more promising in overcoming the training challenges of SNNs compared to existing conversion methods.
However, most of the existing direct training methods only focus on image classification but not RL tasks.

Besides the training methods, there is another challenge in Deep Spiking Reinforcement Learning, i.e., how to distinguish optimal action from highly similar Q-values \cite{Tan}.
It has been proved that, in the process of optimizing a ANN for image classification, the value of the correct class is always significantly higher than the wrong calsses.
In contrast to image classificatoin, the Q-values of different actions are often very similar even for a well-trained network in Reinforcement Learning \cite{Tan}.
Actually, the confusing Q-value issue is not really a problem in Reinforcement Learning, because of the continues information representation of traditional ANNs.
But in Deep Spiking Reinforcement Learning, how to make the discrete spikes outputed by SNNs represent these highly similar Q-values well is a challenging problem.

To address these issues, we propose a Deep Spiking Q-Network (DSQN) in this paper.
Specifically, we use Leaky Integrate-and-Fire (LIF) neurons in DSQN with firing rate coding and appropriate but extremely short simulation time window (64 timesteps) to address the issue of the confusing Q-values.
In addition, we adapt a spiking surrogate gradient learning algorithm to achieve the direct training for DSQN.
Whereafter, we demonstrate the advantages of using LIF neurons in DSQN theoretically.

\begin{figure*}[htbp]
    \centering
    \includegraphics[scale=0.85]{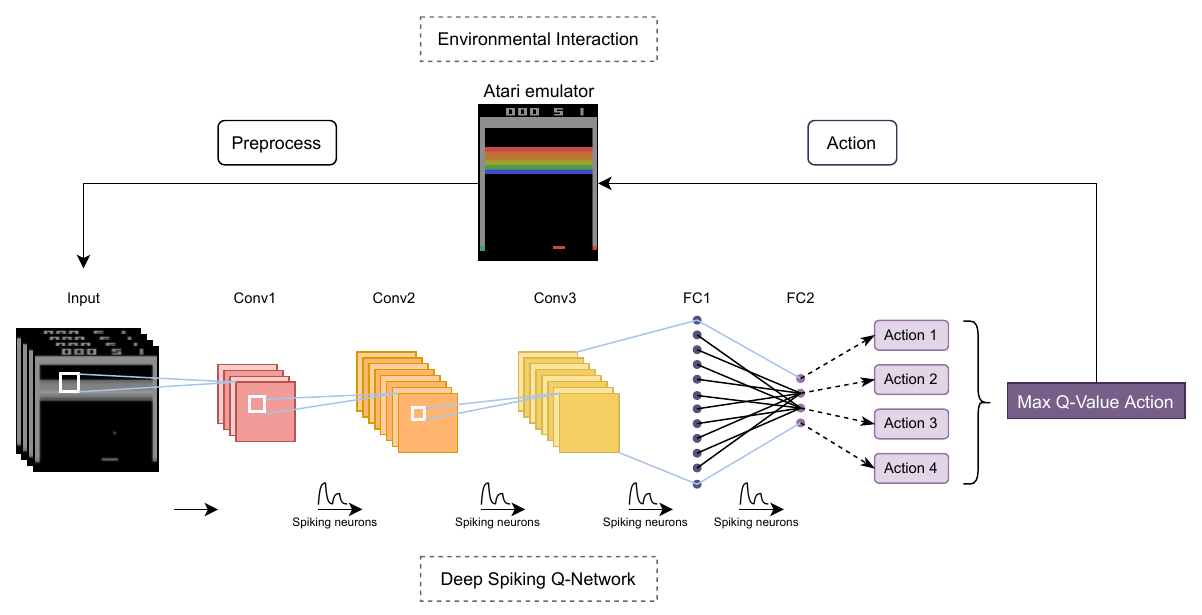}
    \caption{The architecture and environmental interaction of Deep Spiking Q-Network which consists of 3 convolution layers and 2 fully connected layers.}
    \label{fig:dsqn}
\end{figure*}

In the end, comprehensive experiments have been conducted on 17 top-performing Atari games.
And the experimental results show that DSQN completely surpasses the conversion-based SNN \cite{Tan} in terms of performance, stability, generalization, and energy-efficiency.
At the same time, DSQN reaches the same performance level of the vanilla DQN \cite{Mnih}.
Our method provides another way to achieve high performance on Atari games with SNNs while avoiding the limitations of conversion methods.
To the best of our knowledge, our work is the first one to achieve state-of-the-art performance on multiple Atari games with the directly-trained SNN.
And it paves the way for further research on solving Reinforcement Learning problems with directly-trained SNNs.

\section{Related Works}
\cite{Mnih} introduced deep neural networks into the Q-Learning, a traditional RL algorithm, and formed the DQN algorithm, created the field of DRL.
They used convolutional neural networks to approximate the Q function of Q-learning, as the result, DQN achieved and even surpassed human-level on 49 Atari games.
After that, \cite{Devdhar} is the first work to introduce spiking conversion methods to the domain of deep Q-learning.
They demonstrated that both shallow and deep ReLU networks can be converted to SNNs without performance degradation on Atari game Breakout.
Then, they showed that the converted SNN is more robust to input perturbations than the original neural network.
However, it only focuses on improving the robustness rather than the performance of SNNs on Atari games.
To further improve the performance, \cite{Tan} proposed a more effective conversion method based on the more accurate approximation of the spiking firing rates.
It reduced the conversion error based on a pre-trained DQN, and achieved state-of-the-art performance on multiple Atari games.
Although these conversion-based researches have led to further development of DSRL, there are still some limitations that remain unsolved, e.g., the heavy dependence on pre-trained ANNs and demand for very long simulation time window.
Other related works are \cite{Yuan, Guangzhi, Shen, Shen1, Shen2}.

In contrast to the existing methods, our method is directly trained by spiking surrogate gradient learning on LIF neurons.
This makes it more flexible and reduces the training cost because it has no dependence on pre-trained ANNs and only requires extremely short simulation time window.

\section{Methods}
In this section, we describe the Deep Spiking Q-Network (DSQN) in detail, including the directly-trained Deep Spiking Reinforcement Learning architecture, direct learning method, and theoretically demonstration of the advantages of using LIF neurons in DSQN.

\subsection{Architecture of DSQN}
The Deep Spiking Q-Networks consists of \begin{math}3\end{math} convolution layers and \begin{math}2\end{math} fully connected layers.
We use LIF neurons in DSQN to from a directly trained Deep Spiking Reinforcement Learning architecture.
With firing rate encoding and appropriate length of simulation time window, this architecture can achieve sufficient accuracy to handle the confusing Q-value issue mentioned in Sectoin 1, while maintaining the energy-efficiency advantage of SNNs with direct learning method.
Figure \ref{fig:dsqn} concretely shows the architecture and environmental interaction of DSQN.

For a network with \begin{math}L\end{math} layers, let \begin{math}V ^{l, t}\end{math} denote the membrane potential of neurons in layer \begin{math}l\end{math} at simulation time \begin{math}t\end{math}.
The LIF neuron integrates inputs until the membrane potential exceeds a threshold \begin{math}V _{\text{th}} \in \mathbb{R} ^{+}\end{math}, and a spike correspondingly generated.
Once the spike is generated, the membrane potential will be reset by \textit{hard reset} or \textit{soft reset}.
Note that the \textit{hard reset} means resetting the membrane potential back to a baseline, typically \begin{math}0\end{math}.
The \textit{soft reset} means subtracting the threshold \begin{math}V _{\text{th}}\end{math} from the membrane potential when it exceeds the threshold.

The neuronal dynamics of the LIF neurons in layer \begin{math}l \in \{1, \cdots, L - 1\}\end{math} at simulation time \begin{math}t\end{math} could be described as follows:

\begin{equation}
    U ^{l, t} = V ^{l, t - 1} + \frac{1}{\tau _{\text{m}}} (W ^{l} S ^{l - 1, t} - V ^{l, t - 1} + V _{\text{r}}),
    \label{eq:LIFu}
\end{equation}

\begin{equation}
    V ^{l, t} =   \begin{cases}
                        U ^{l, t} (1 - S ^{l, t}) + V _{\text{r}} S ^{l, t} & \textit{hard reset} \\
                        U ^{l, t} - V _{\text{th}} S ^{l, t} & \textit{soft reset}
                    \end{cases}.
    \label{eq:LIFv}
\end{equation}

\noindent Equation \eqref{eq:LIFu} describes the sub-threshold membrane potential of neurons, that is, when the membrane potential does not exceed the threshold potential \begin{math}V _{\text{th}}\end{math}, in which \begin{math}\tau _{\text{m}}\end{math} denotes the membrane time constant, \begin{math}W ^{l}\end{math} denotes the learnable weights of the neurons in layer \begin{math}l\end{math}, \begin{math}V _{\text{r}}\end{math} denotes the initial membrane potential.
Equation \eqref{eq:LIFv} describes the membrane potential of neurons when reached \begin{math}V _{\text{th}}\end{math}.

The output of the LIF neurons in layer \begin{math}l \in \{1, \cdots, L - 1\}\end{math} at simulation time \begin{math}t\end{math} could be expressed as follows:

\begin{equation}
     S ^{l, t} = \Theta (U ^{l, t} - V _{\text{th}}),
    \label{eq:spikeOutput}
\end{equation}

\begin{equation}
    \Theta (x) =    \begin{cases}
                        1 & x \geq 0 \\
                        0 & \text{otherwise}
                    \end{cases},
    \label{eq:SpikingFunction}
\end{equation}

\noindent where \begin{math}\Theta (x)\end{math} is the spiking function of neurons.

\begin{figure}[htbp]
    \centering
    \includegraphics[scale=0.55]{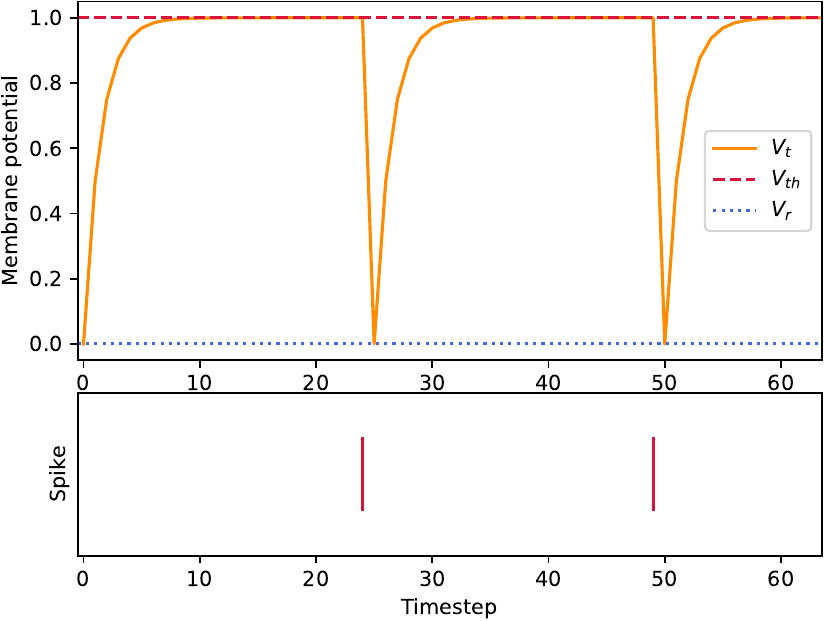}
    \caption{The neuronal dynamics of the LIF neuron which resets by \textit{hard reset} when the input is constant at \begin{math}1\end{math} for the length of simulation time window \begin{math}t = 64\end{math}, the membrane time constant \begin{math}\tau _{\text{m}} = 2\end{math}, the initial membrane potential \begin{math}V _{\text{r}} = 0\end{math}, and the threshold potential \begin{math}V_{th} = 1\end{math}.}
    \label{fig:LIF}
\end{figure}

The neuronal dynamics of the LIF neuron which is reset by \textit{hard reset} is illustrated in Figure \ref{fig:LIF}.
As the simulation time passes, the LIF neuron integrates the input current, and its membrane potential continues to rise according to Equation \eqref{eq:LIFu}.
Until the membrane potential reaches the membrane potential threshold \begin{math}V _{\text{th}}\end{math}, a spike is emitted by the LIF neuron according to Equation \eqref{eq:spikeOutput} and \eqref{eq:SpikingFunction}.
Then the membrane potential is reset according to Equation \eqref{eq:LIFv}.

For the neurons in the final layer \begin{math}L\end{math}, their output \begin{math}O ^{L}\end{math} could be described by

\begin{equation}
    O ^{L} = W ^{L} \frac{1}{t} \sum _{t ^{\prime} = 1} ^{t} S ^{L - 1, t ^{\prime}},
    \label{eq:Qoutput}
\end{equation}

\noindent where \begin{math}W ^{L}\end{math}is the learnable weights of the neurons in the final layer.
At the same time, \begin{math}O ^{L}\end{math} denotes the output Q-values of DSQN.

As a RL agent, during the interacting with the environment, DSQN is trained by deep Q-Learning algorithm \cite{Mnih}.
Thus, the deep spiking Q-learning algorithm uses the following loss function:

\begin{equation}
    \cal{L}(W) = \mathbb{E} _{(s, a, r, s ^{\prime}) \sim U(D)} \left [(y _{(r, s ^{\prime})} - Q(s, a; W)) ^{2} \right ],
    \label{eq:loss}
\end{equation}

\noindent with

\begin{equation}
    y _{(r, s ^{\prime})} = r + \gamma \ \underset{a ^{\prime}}{\text{max}} \ Q(s ^{\prime}, a ^{\prime}; W ^{-}),
    \label{eq:y}
\end{equation}

\begin{figure*}[htbp]
    \centering
    \subfloat[\begin{footnotesize}Arc-tangent function with \begin{math}\alpha = 2\end{math}\end{footnotesize}]{
        \includegraphics[scale=0.45]{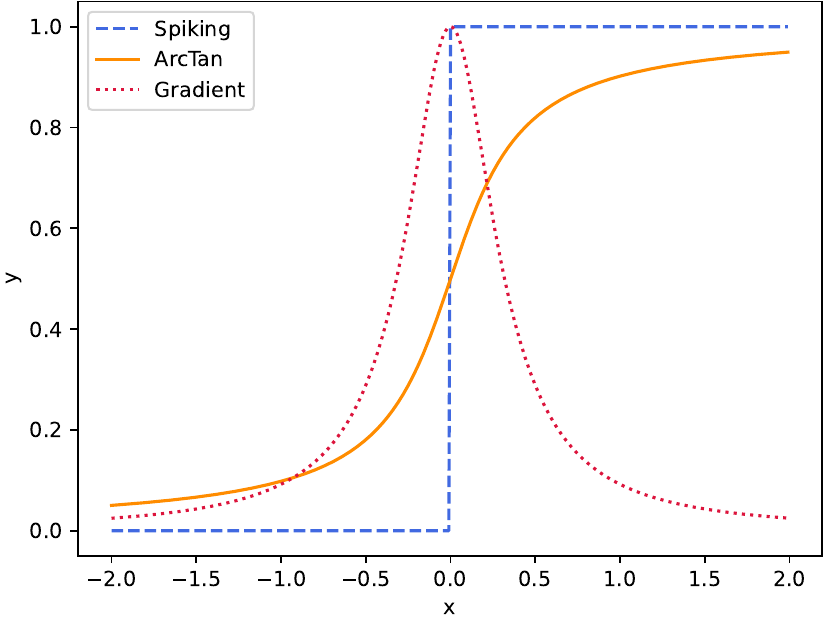}
        \label{subfig:sgl2}
    }
    \hspace{0.5 in}
    \subfloat[\begin{footnotesize}Arc-tangent function with \begin{math}\alpha = 4\end{math}\end{footnotesize}]{
        \includegraphics[scale=0.45]{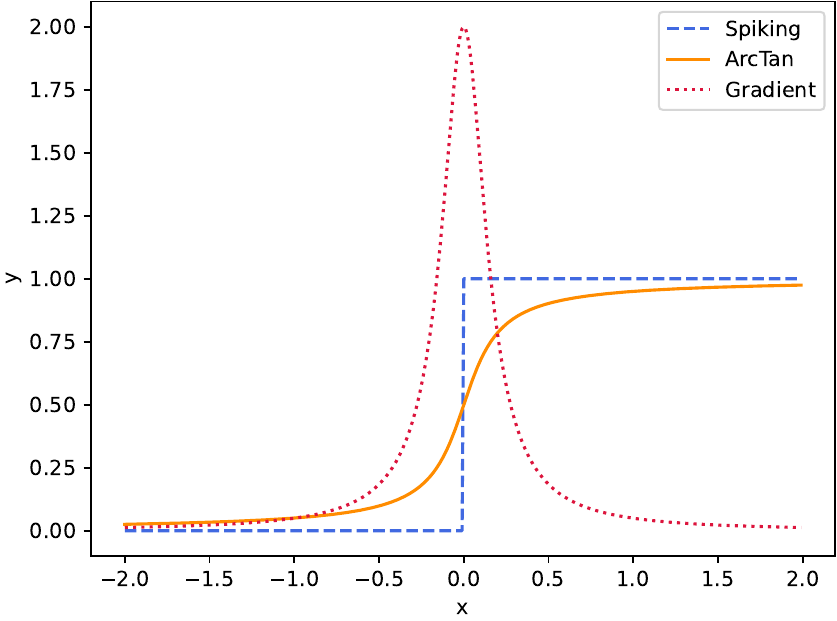}
        \label{subfig:sgl4}
    }
    \\
    \subfloat[\begin{footnotesize}Sigmoid function with \begin{math}\alpha = 2\end{math}\end{footnotesize}]{
        \includegraphics[scale=0.45]{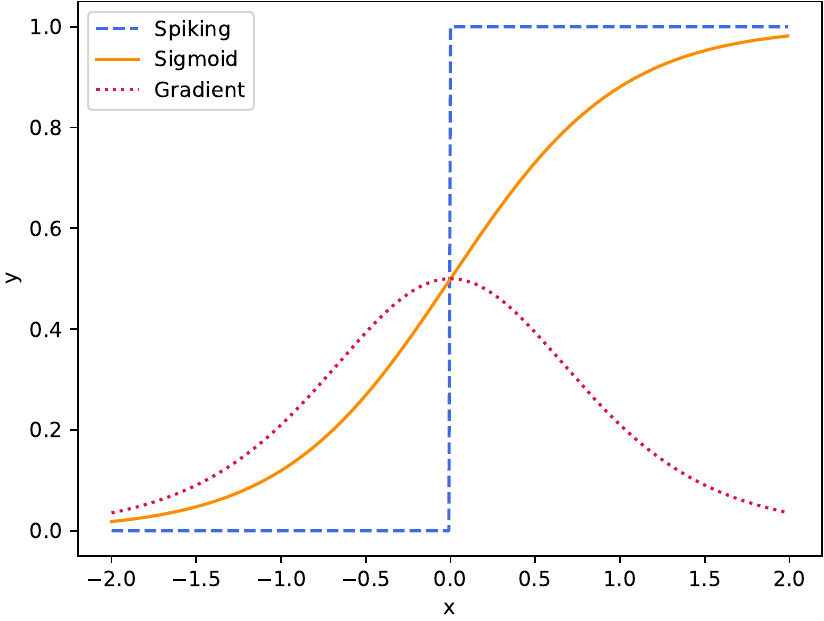}
        \label{subfig:sigmoid2}
    }
    \hspace{0.5 in}
    \subfloat[\begin{footnotesize}Sigmoid function with \begin{math}\alpha = 4\end{math}\end{footnotesize}]{
        \includegraphics[scale=0.45]{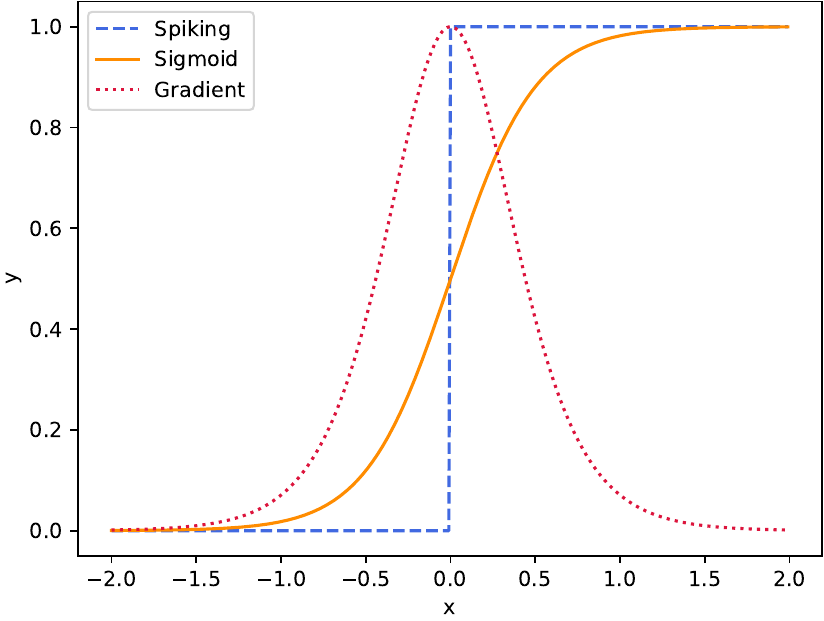}
        \label{subfig:sigmoid4}
    }
    \caption{The curves of the spiking function \begin{math}\Theta (x)\end{math}, the two commonly used surrogate gradient functions \begin{math}\sigma _{\text{arctan}} (\alpha x), \sigma _{\text{sigmoid}} (\alpha x)\end{math}, and their gradient functions \begin{math}\sigma _{\text{arctan}} ^{\prime} (\alpha x), \sigma _{\text{sigmoid}} ^{\prime} (\alpha x)\end{math} with different \begin{math}\alpha\end{math}.}
    \label{fig:SGL}
\end{figure*}

\noindent where \begin{math}Q(s, a; W)\end{math} denotes the approximate Q-value function parameterized by DSQN, \begin{math}W\end{math} and \begin{math}W ^{-}\end{math} denote the weights of DSQN at the current and history, respectively.
\begin{math}(s, a, r, s ^{\prime}) \sim U(D)\end{math} denotes the minibatches drawn uniformly at random from the experience replay memory \begin{math}D\end{math}, and \begin{math}\gamma\end{math} is the reward discount factor.

According to Equation \eqref{eq:Qoutput}, the loss function could also be simply expressed by

\begin{equation}
    \cal{L}(W) = \mathbb{E} \left [(y - O ^{L}) ^{2} \right ].
    \label{eq:loss1}
\end{equation}

\subsection{Direct learning method for DSQN}
In this section, we only consider the case of LIF neurons which are reset by \textit{hard reset}.

According to Equation \eqref{eq:Qoutput}, \eqref{eq:loss1} and the chain rule, for the final layer \begin{math}L\end{math}, we have

\begin{equation}
    \frac{\partial \cal{L}}{\partial W ^{L}} = \frac{2}{t} \mathbb{E} \left [O ^{L} - y \right ] \sum _{t ^{\prime} = 1} ^{t} S ^{L - 1, t ^{\prime}} ,
    \label{eq:BPoutputLayer}
\end{equation}

\noindent and for the hidden layers \begin{math}l \in \{1, \cdots, L - 1\}\end{math}, according to Equation \eqref{eq:LIFu}, \eqref{eq:spikeOutput} and \eqref{eq:loss1}, we have

\begin{equation}
        \frac{\partial \cal{L}}{\partial W ^{l}} = \sum _{t ^{\prime} = 1} ^{t} \frac{\partial \cal{L}}{\partial S ^{l, t ^{\prime}}} \frac{\partial S ^{l, t ^{\prime}}}{\partial U ^{l, t ^{\prime}}} \frac{\partial U ^{l, t ^{\prime}}}{\partial W ^{l}}.
    \label{eq:BPhiddenLayer}
\end{equation}

The first factor in Equation \eqref{eq:BPhiddenLayer} could be derived as

\begin{equation}
    \begin{split}
        \frac{\partial \cal{L}}{\partial S ^{l, t ^{\prime}}} & = \frac{\partial \cal{L}}{\partial S ^{l + 1, t ^{\prime}}} \frac{\partial S ^{l + 1, t ^{\prime}}}{\partial U ^{l + 1, t ^{\prime}}} \frac{\partial U ^{l + 1, t ^{\prime}}}{\partial S ^{l, t ^{\prime}}} \\
                                                                            & = \frac{\partial \cal{L}}{\partial S ^{l + 1, t ^{\prime}}} \frac{\partial \Theta (U ^{l + 1, t ^{\prime}} - V _{\text{th}})}{\partial U ^{l + 1, t ^{\prime}}} \frac{W ^{l + 1}}{\tau _{\text{m}}}.
    \end{split}
    \label{eq:BPfirstFactor}
\end{equation}

Specially, when \begin{math}l = L - 1\end{math}, Equation \eqref{eq:BPfirstFactor} is different:

\begin{equation}
    \begin{split}
        \frac{\partial \cal{L}}{\partial S ^{l, t ^{\prime}}} & = \frac{\partial \cal{L}}{\partial O ^{L}} \frac{\partial O ^{L}}{\partial S ^{l, t ^{\prime}}} \\
                                                                            & = \frac{2}{t} W ^{l} \mathbb{E} \left [O ^{L} - y \right ].
    \end{split}
    \label{eq:BPfirstFactor1}
\end{equation}

The spiking function \begin{math}\Theta (x)\end{math} is non-differentiable, this leads to that Equation \eqref{eq:BPhiddenLayer} which is also non-differentiable.
To address this issue, we use a differentiable surrogate gradient function \cite{Neftci} \begin{math}\sigma (x)\end{math} to approximate \begin{math}\Theta (x)\end{math}.
Correspondingly, the Equation \eqref{eq:BPfirstFactor} can be rewritten as:

\begin{equation}
        \frac{\partial \cal{L}}{\partial S ^{l, t ^{\prime}}} = \frac{\partial \cal{L}}{\partial S ^{l + 1, t ^{\prime}}} \frac{\partial \sigma (U ^{l + 1, t ^{\prime}} - V _{\text{th}})}{\partial U ^{l + 1, t ^{\prime}}} \frac{W ^{l + 1}}{\tau _{\text{m}}}.
    \label{eq:BPfirstFactorSG}
\end{equation}

The third factor in Equation \eqref{eq:BPhiddenLayer} could be derived as:

\begin{equation}
    \begin{split}
        \frac{\partial U ^{l, t ^{\prime}}}{\partial W ^{l}}    & = \frac{\partial U ^{l, t ^{\prime}}}{\partial V ^{l, t ^{\prime} - 1}} \frac{\partial V ^{l, t ^{\prime} - 1}}{\partial W ^{l}} + \frac{1}{\tau _{\text{m}}} S ^{l - 1, t ^{\prime}} \\
                                                                & = M ^{l, t ^{\prime}} \frac{\partial U ^{l, t ^{\prime} - 1}}{\partial W ^{l}} + \frac{S ^{l - 1, t ^{\prime}}}{\tau _{\text{m}}},
    \end{split}
     \label{eq:BPthirdFactor}
\end{equation}

\noindent where

\begin{small}
    \begin{equation}
        M ^{l, t} = (1 - \frac{1}{\tau _{\text{m}}}) \left [1 - S ^{l, t - 1} + \frac{\partial \sigma (U ^{l, t - 1} - V _{\text{th}})}{\partial U ^{l, t - 1}}(V _{\text{r}} - U ^{l, t - 1}) \right ].
        \label{eq:M}
    \end{equation}
\end{small}

\noindent Then, we continuous derive gradients across time dimension.
When \begin{math}t = 1\end{math}, we have

\begin{equation}
    \frac{\partial U ^{l, t ^{\prime}}}{\partial W ^{l}} = \frac{S ^{l - 1, 1}}{\tau _{\text{m}}},
    \label{eq:BPthirdFactor1}
\end{equation}

\noindent when \begin{math}t > 1\end{math}, we obtain:

\begin{equation}
    \frac{\partial U ^{l, t ^{\prime}}}{\partial W ^{l}} = \prod _{\tau = 1} ^{t ^{\prime}} M ^{l, \tau} + \sum _{\tau = 1} ^{t ^{\prime} - 1} \prod _{i = \tau} ^{t ^{\prime}} M ^{l, i} \frac{S ^{l - 1, \tau}}{\tau _{\text{m}}} + \frac{S ^{l - 1, t ^{\prime}}}{\tau _{\text{m}}}.
    \label{eq:BPthirdFactor2}
\end{equation}

The commonly used surrogate gradient functions are arc-tangent function and sigmoid function, both of them are very similar to \begin{math}\Theta (x)\end{math}.
In this paper, considering that the complexity of surrogate gradient function would affect the computational efficiency, thus, we use arc-tangent function in DSQN, which is defined by

\begin{equation}
    \sigma _{\text{arctan}} (\alpha x) = \frac{1}{\pi} arctan(\frac{\pi}{2} \alpha x) + \frac{1}{2},
    \label{eq:ATan}
\end{equation}

\noindent where \begin{math}\alpha\end{math} is the factor that controls the smoothness of the function.
The gradient of arc-tangent function could be expressed as

\begin{equation}
    \sigma _{\text{arctan}} ^{\prime} (\alpha x) = \frac{\alpha}{2 \left [1 + (\frac{\pi}{2} \alpha x) ^{2} \right ]}.
    \label{eq:ATanGradient}
\end{equation}

Figure \ref{fig:SGL} shows the curves of the spiking function \begin{math}\Theta (x)\end{math}, the two commonly used surrogate gradient functions \begin{math}\sigma _{\text{arctan}} (\alpha x), \sigma _{\text{sigmoid}} (\alpha x)\end{math}, and their gradient functions \begin{math}\sigma _{\text{arctan}} ^{\prime} (\alpha x), \sigma _{\text{sigmoid}} ^{\prime} (\alpha x)\end{math} with different \begin{math}\alpha\end{math}.
Notably, the larger \begin{math}\alpha\end{math} is, the closer \begin{math}\sigma _{\text{arctan}} (\alpha x)\end{math} and \begin{math}\Theta (x)\end{math} will be.
Meanwhile, when \begin{math}x\end{math} is near \begin{math}0\end{math}, the gradient will be more likely to explode, and when \begin{math}x\end{math} moves away from \begin{math}0\end{math}, the gradient will be more likely to disappear.

Using surrogate gradient function allows DSQN to be directly trained well by deep spiking Q-learning algorithm.

\subsection{Demonstration of using LIF neurons in DSQN}
In this section, we demonstrate the advantages of using LIF neurons in directly-trained Deep Spiking Q-Networks theoretically.
We first explain why existing conversion methods in Deep Spiking Reinforcement Learning use Integrate-and-Fire (IF) neurons, then provide the theory for using LIF neurons in directly-trained DSQN.

\subsubsection{Limitation of conversion methods in DSRL}
\cite{Rueckauer} demonstrated the IF neuron which is reset by \textit{soft reset} is an unbiased estimator of ReLU activation function over time.
The neuronal dynamics of the IF neuron could be described as

\begin{equation}
    V ^{l, t} = \begin{cases}
                    (V ^{l, t - 1} + z ^{l, t})(1 - S ^{l, t}) + V _{\text{r}} S ^{l, t} & \textit{hard reset} \\
                    V ^{l, t - 1} + z ^{l, t} - V _{\text{th}} S ^{l, t} & \textit{soft reset}
                \end{cases},
    \label{eq:IFv}
\end{equation}

With \begin{math}V _{\text{r}} = 0\end{math} and the fact that the input of the first layer \begin{math}z ^{1} = V _{\text{th}} a ^{1}\end{math} constantly, \cite{Rueckauer} provided the relationship between the firing rate \begin{math}r ^{1, t}\end{math} of IF neurons in the first layer and the output of ReLU activation function \begin{math}a ^{1}\end{math} when receiving the same inputs as:

\begin{figure}[htbp]
    \centering
    \includegraphics[scale=0.55]{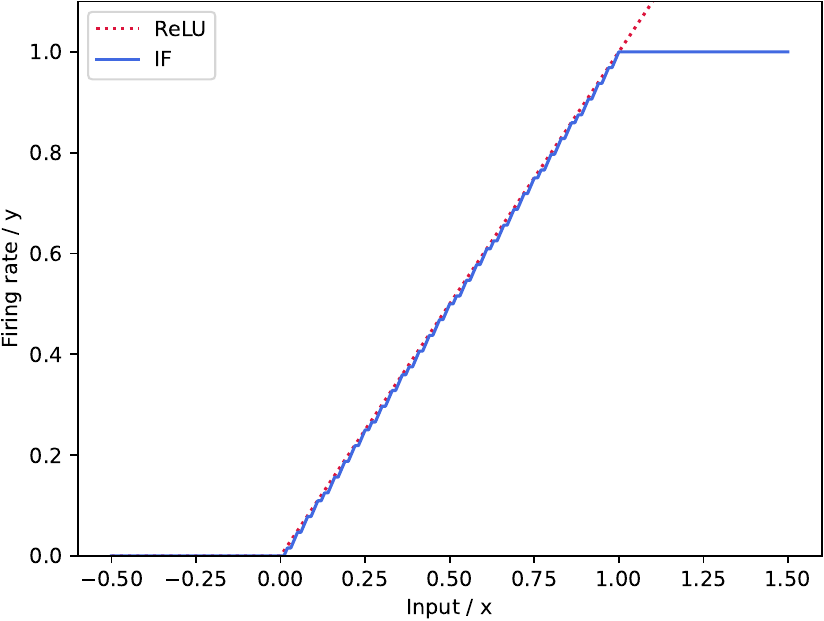}
    \caption{The firing rate of the IF neuron which is reset by \textit{soft reset} and the output of ReLU function.}
    \label{fig:reluAndIF}
\end{figure}

\begin{equation}
    r ^{1, t} = \begin{cases}
                    a ^{1} r _{\text{max}} \frac{V _{\text{th}}}{V _{\text{th}} + \epsilon ^{1}} - \frac{V ^{1, t}}{t (V _{\text{th}} + \epsilon ^{l})} & \textit{hard reset} \\
                    a ^{1} r _{\text{max}} - \frac{V ^{1, t}}{t V _{\text{th}}} & \textit{soft reset}
                \end{cases},
    \label{eq:IFfiringRate}
\end{equation}

\noindent where \begin{math}r _{\text{max}}\end{math} denotes the maximum firing rate, and \begin{math}\epsilon\end{math} denotes the residual charge which is discarded at resetting.

According to Equation \eqref{eq:IFfiringRate}, when the length of simulation time window \begin{math}t\end{math} tends to infinity and the inputs are in \begin{math}[0, 1]\end{math}, the IF neuron which is reset by \textit{soft reset} is an unbiased estimator of ReLU activation function over time.

Figure \ref{fig:reluAndIF} shows the relationship between the firing rate of the IF neuron which is reset by \textit{soft reset} and the output of ReLU activation function when receiving the same inputs.
When \begin{math}x \in [0, 1]\end{math}, the firing rate of the IF neuron is highly similar to the output of ReLU activation function.
But when input \begin{math}x \geq 1\end{math}, the firing rate of the IF neuron no longer increase, because \begin{math}r _{\text{max}} = 1\end{math}.

Therefore, the existing conversion methods in DSRL must introduce normalization technique to normalize the inputs beyond \begin{math}1\end{math} into \begin{math}[0, 1]\end{math}, and then ANNs can be successfully converted to SNNs.
This results in the existing conversion methods in DSRL requiring very long simulation time window, typically hundreds timesteps, to achieve sufficient accuracy, so as to minimize the error generated during the converting process.
In addition, the inherent problem of the conversion methods, i.e., the heavy dependence on pre-trained ANNs, remains unsolved.

\subsubsection{Advantages of using LIF neurons in DSQN}
Similar to the process of deriving Equation \eqref{eq:IFfiringRate}, we could derive an equation describing the firing rate of the LIF neurons in the first layer from Equation \eqref{eq:LIFu} and \eqref{eq:LIFv}.
To simplify the notation, we drop the layer and neuron indices, let \begin{math}z\end{math} denote the input and \begin{math}V _{\text{r}} = 0\end{math}.

For \textit{hard reset}, starting from Equation \eqref{eq:LIFv}, the average firing rate could be simply computed by summing over the simulation time \begin{math}t\end{math} as

\begin{equation}
    \sum _{t ^{\prime} = 1} ^{t} V ^{t ^{\prime}} = \frac{1}{\tau _{\text{m}}} \sum _{t ^{\prime} = 1} ^{t} \left [(\tau _{\text{m}} - 1) V ^{t ^{\prime} - 1} + z) \right ](1 - S ^{t ^{\prime}}).
    \label{eq:S1}
\end{equation}

\noindent Under the assumption of constant input \begin{math}z\end{math} to the first layer, and using \begin{math}N ^{t} = \sum _{t ^{\prime} = 1} ^{t} S ^{t ^{\prime}}\end{math}, we obtain:

\begin{equation}
    \tau _{\text{m}} \sum _{t ^{\prime} = 1} ^{t} V ^{t ^{\prime}} = (\tau _{\text{m}} - 1) \sum _{t ^{\prime} = 1} ^{t} V ^{t ^{\prime} - 1} (1 - S ^{t ^{\prime}}) + z (\frac{t}{\Delta t} - N ^{t}).
    \label{eq:S2}
\end{equation}

\begin{figure*}[htbp]
    \centering
    \subfloat[\begin{footnotesize}The IF neuron which is reset by \textit{hard reset}.\end{footnotesize}]{
        \includegraphics[scale=0.45]{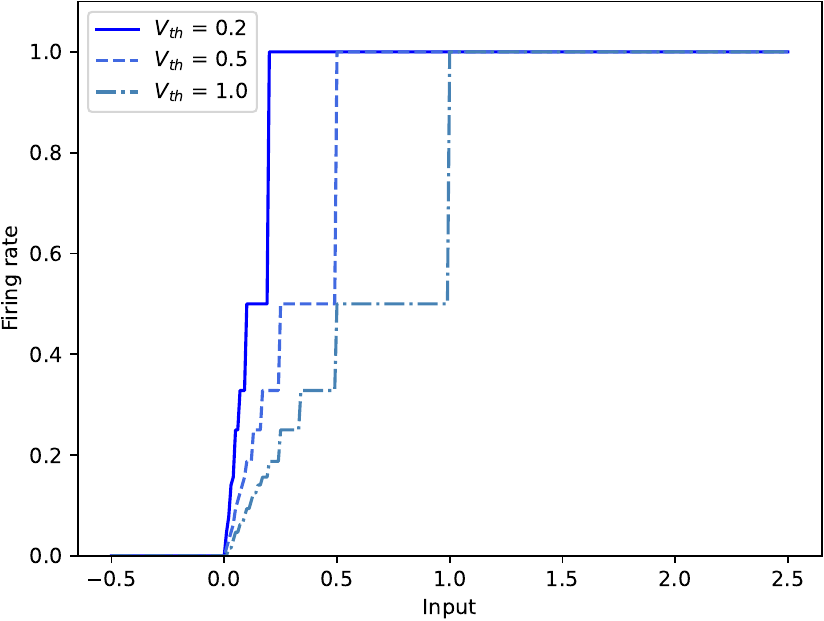}
        \label{subfig:ifFR0}
    }
    \hspace{0.5 in}
    \subfloat[\begin{footnotesize}The IF neuron which is reset by \textit{soft reset}.\end{footnotesize}]{
        \includegraphics[scale=0.45]{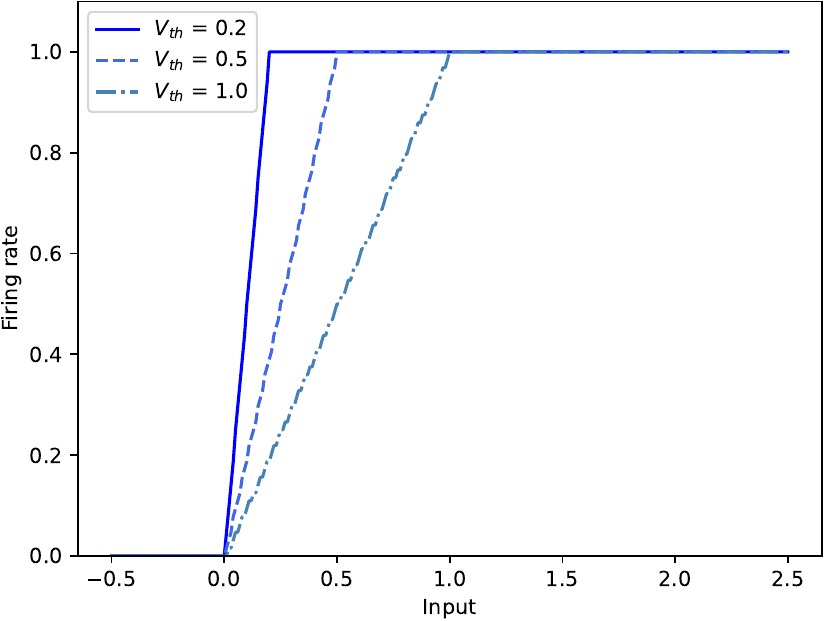}
        \label{subfig:ifFRnone}
    }
    \\
    \subfloat[\begin{footnotesize}The LIF neuron which is reset by \textit{hard reset}.\end{footnotesize}]{
        \includegraphics[scale=0.45]{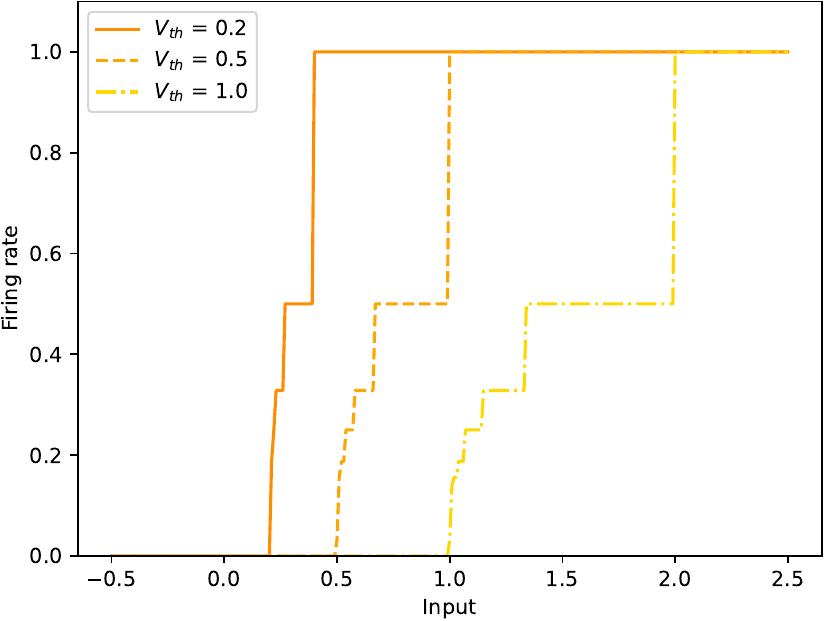}
        \label{subfig:lifFR20}
    }
    \hspace{0.5 in}
    \subfloat[\begin{footnotesize}The LIF neuron which is reset by \textit{soft reset}.\end{footnotesize}]{
        \includegraphics[scale=0.45]{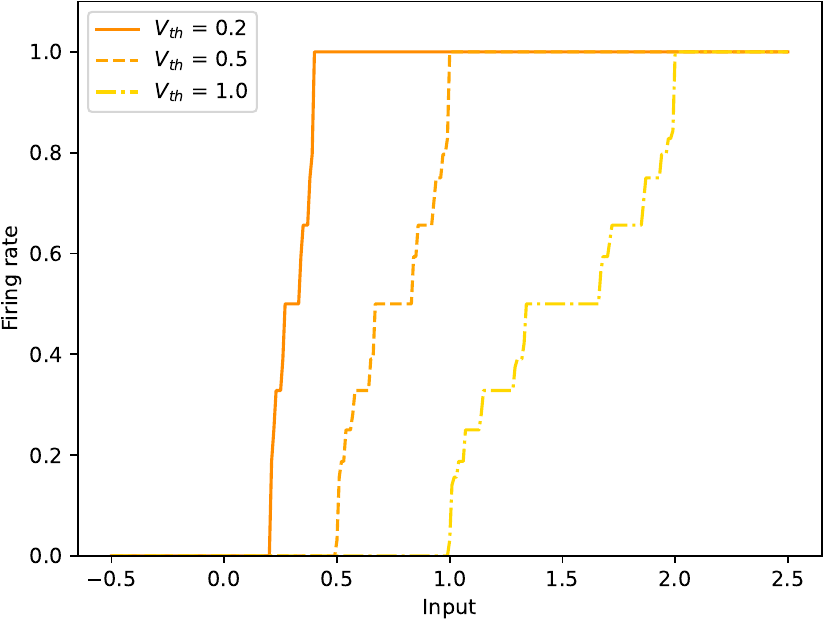}
        \label{subfig:lifFR2none}
    }
    \caption{The relationship between the firing rate and inputs of the IF neuron and LIF neuron which are reset by \textit{hard reset} and \textit{soft reset}.
            \begin{math}V _{\text{th}}\end{math} is set to \begin{math}0.2\end{math}, \begin{math}0.5\end{math}, \begin{math}1\end{math}, respectively.
            And \begin{math}V _{\text{r}} = 0\end{math} constantly.}
    \label{fig:IFandLIF}
\end{figure*}

\noindent The time resolution \begin{math}\Delta t\end{math} enters Equation \eqref{eq:S2} when evaluating the time-sum over a constant: \begin{math}\sum _{t ^{\prime} = 1} ^{t} 1 = \frac{t}{\Delta t}\end{math}.
It will be replaced by the definition of the maximum firing rate \begin{math}r _{\text{max}} = \frac{1}{\Delta t}\end{math} in the following.

\noindent After rearranging Equation \eqref{eq:S2} to yield the total number of spikes \begin{math}N ^{t}\end{math} at simulation time \begin{math}t\end{math}, dividing by the simulation time \begin{math}t\end{math}, setting \begin{math}V ^{0} = 0\end{math}, and reintroducing the dropped indices, we obtain the average firing rate \begin{math}r\end{math} of the LIF neurons in the first layer:

\begin{small}
    \begin{equation}
        r ^{1, t} = r _{\text{max}} - \frac{1}{t z ^{1, t}} \{ \tau _{\text{m}} V ^{1, t} + \sum _{t ^{\prime} = 1} ^{t} [V ^{1, t ^{\prime} - 1} + (\tau _{\text{m}} - 1) V ^{1, t ^{\prime} - 1} S ^{1, t ^{\prime}}] \}.
        \label{eq:LIFfiringRateH}
    \end{equation}
\end{small}

For \textit{soft reset}, averaging Equation \eqref{eq:LIFv} over the simulation time \begin{math}t\end{math} yields:

\begin{equation}
    \frac{1}{t} \sum _{t ^{\prime} = 1} ^{t} V ^{t ^{\prime}} = \frac{\tau _{\text{m}} - 1}{t \tau _{\text{m}}} \sum _{t ^{\prime} = 1} ^{t} V ^{t ^{\prime} - 1} + \frac{z}{\tau _{\text{m}}} r _{\text{max}} - V _{\text{th}} \frac{N ^{t}}{t}.
    \label{eq:S3}
\end{equation}

\noindent Using \begin{math}z ^{1} = V _{\text{th}} a ^{1}\end{math}, solving for \begin{math}r  = N / t\end{math}, setting \begin{math}V ^{0} = 0\end{math} and reintroducing the indices yields:

\begin{equation}
    r ^{1, t} = \frac{1}{\tau _{\text{m}}} (a ^{1} r _{\text{max}} - \frac{1}{t V _{\text{th}}} \sum _{t ^{\prime} = 1} ^{t} V ^{1, t ^{\prime} - 1}) - \frac{V ^{1, t}}{t V _{\text{th}}}.
    \label{eq:LIFfiringRateS}
\end{equation}

Figure \ref{fig:IFandLIF} shows the relationship between the firing rate and inputs of the IF and LIF neurons which are reset by \textit{hard reset} and \textit{soft reset} with different \begin{math}V _{\text{th}}\end{math}.

In practice, only one particular threshold could be used in the training and testing stage, and the ranges of the neuron inputs with fixed threshold are similar.
However, in the hyperparameter tuning stage, the LIF neuron provides wider ranges for different thresholds.
Thus, LIF neurons have more potential to get optimal results in our DSQN and could be directly trained without relying on the normalization technique in conversion methods.

\section{Experimental Results}

In this section, we evaluated the Deep Spiking Q-Network on 17 top-performing Atari games.
Then, we compared the experimental results of DSQN and the conversion-based SNN \cite{Tan} using the vanilla DQN \cite{Mnih} as a benchmark, and analyzed the experimental results in several aspects.

\subsection{Experimental setup}
\subsubsection{Environments}
We performed the experiments based on OpenAI Gym\footnote{The open-source code of OpenAI Gym could be accessed at \url{https://github.com/openai/gym}.}.
Each experiment ran on a single GPU.
The specific experimental hardware environments is shown in Table \ref{tab:hardware}.

\begin{table}[htbp]
    \centering
    \caption{The experimental hardware environment.}
    \scalebox{1.0}{
        \begin{tabular}{cc}
            \textbf{Item}   & \textbf{Detail} \\
            \toprule
            CPU             & Intel Xeon Silver 4116 \\
            GPU             & NVIDIA GeForce RTX 2080Ti \\
            OS              & Ubuntu 16.04 LTS \\
            Memory          & At least \SI{40}{GB} for a single experiment \\
            \bottomrule
        \end{tabular}
    }
    \label{tab:hardware}
\end{table}

\subsubsection{Parameters}
The proposed DSQN consists of three convolution layers and two fully connected layers.
The specific parameters of the three convolution layers are shown in Table \ref{tab:convP}, while the two fully connected layers use different paramaters.
The first fully connected layer FC1 has 512 neurons, and the final layer FC2 has different neurons on different Atari games, from 4 to 18, depending on the number of validate actions in the game.

\begin{table}[htbp]
    \centering
    \caption{The parameters of the convolution layers in DSQN.}
    \scalebox{1.0}{
        \begin{tabular}{cccc}
            \textbf{Layer}  & \textbf{Number of kernels}    & \textbf{Kernel size}              & \textbf{Stride} \\
            \toprule
            Conv1           & \begin{math}32\end{math}      & \begin{math}8 \times 8\end{math}  & \begin{math}4\end{math} \\
            Conv2           & \begin{math}64\end{math}      & \begin{math}4 \times 4\end{math}  & \begin{math}2\end{math} \\
            Conv3           & \begin{math}64\end{math}      & \begin{math}3 \times 3\end{math}  & \begin{math}1\end{math} \\
            \bottomrule
        \end{tabular}
    }
    \label{tab:convP}
\end{table}

Both DSQN and the vanilla DQN were trained for \begin{math}50M\end{math} timesteps with the same architecture and hyperparameters \cite{Mnih}.
The lengths of simulation time window of the conversion-based SNN and DSQN were set to \begin{math}\bm{500}\end{math} and \begin{math}\bm{64}\end{math} timesteps, respectively.
Other DSQN-specific hyperparameters are illustrated in Table \ref{tab:HP}.

\begin{table}[htbp]
    \centering
    \caption{The values and descriptions of the DSQN-specific hyperparameters.}
    \scalebox{1.0}{
        \begin{tabular}{ccc}
            \textbf{Hyperparameter}                 & \textbf{Value}                & \textbf{Description}\\
            \toprule
            \begin{math}t\end{math}                 & \begin{math}64\end{math}      & Simulation time window \\
            \begin{math}\tau _{\text{m}}\end{math}  & \begin{math}2\end{math}       & Membrane time constant \\
            \begin{math}V _{\text{th}}\end{math}    & \begin{math}1\end{math}       & Membrane potential threshold \\
            \begin{math}V _{\text{r}}\end{math}     & \begin{math}0\end{math}       & Initial membrane potential \\
            \begin{math}\alpha\end{math}            & \begin{math}2\end{math}       & Smootheness factor \\
            lr                                      & \begin{math}0.0001\end{math}  & Learning rate \\
            \bottomrule
        \end{tabular}
    }
    \label{tab:HP}
\end{table}

It is important to point out that, different from the results reported in \cite{Tan} which were conducted with the best \textit{percentile} in \begin{math}[99.9, 99.99]\end{math}, we set the \textit{percentile} to \begin{math}99.9\end{math} constantly when reproducing the conversion-based SNN.

\subsubsection{Preprocessing}

\begin{figure}[htbp]
    \centering
    \subfloat[\begin{footnotesize}The raw image of Atari game Breakout.\end{footnotesize}]{
        \includegraphics[scale=0.7]{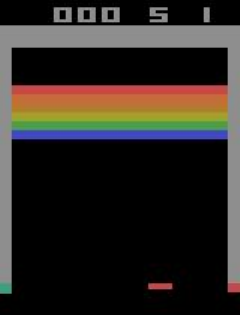}
        \label{subfig:breakoutImage}
    }
    \hspace{0.3 in}
    \subfloat[\begin{footnotesize}The state produced by preprocessing.\end{footnotesize}]{
        \includegraphics[scale=1.1]{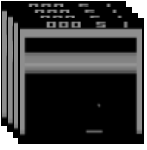}
        \label{subfig:preprocessedImage}
    }
    \caption{Preprocess raw Atari images and stack the \begin{math}m\end{math} most recent frames as a state.}
    \label{fig:Preprocessing}
\end{figure}

Following \cite{Mnih}, to reduce the amount of computation and memory required for training, we preprocess a single raw Atari game frame and stack the \begin{math}m\end{math} most recent frames to produce the input of DSQN, in which \begin{math}m = 4\end{math}.
Figure \ref{fig:Preprocessing} shows the raw Atari image and preprocessed images which were stacked as a state.
Same as the conversion-based SNN and vanilla DQN, we directly use raw real-valued pixel images as the inputs of the neural network without any neural encoding method.

\subsubsection{Testing}
To demonstrate the superiority of DSQN, we choose 17 top-performing Atari games as same as \cite{Tan}, which are also the games where the vanilla DQN has achieved very high performances \cite{Mnih}.
In order to fairly compare DSQN with the conversion-based SNN and vanilla DQN, we reproduced these two methods.

We evaluated these three RL agents by playing 30 rounds at each game with \begin{math}\varepsilon\end{math}-greedy policy (\begin{math}\varepsilon = 0.05\end{math}).
For each round, agents start with different initial random conditions by taking random times (at most 30 times) of \textit{no-op} action, and play for up to 5 minutes (\begin{math}18000\end{math} timesteps).
The mean and standard deviation of the scores obtained in 30 rounds were used as the final scores and standard deviations of these three RL agents.

\subsubsection{Metrics}
We evaluated the performance and stability of the three RL agents over multiple Atari games by the scores and standard deviations they obtained, respectively.
We normalized the scores of DSQN and the conversion-based SNN by the scores of the vanilla DQN.
The standard deviations of the three RL agents were first normalized by their corresponding scores, and then the normalized standard deviations of DSQN and the conversion-based SNN were again normalized by the standard deviations of the vanilla DQN.

\begin{table}[htbp]
    \centering
    \caption{The metrics for normalized scores and standard deviations.}
    \scalebox{1.0}{
        \begin{tabular}{cccc}
            ~                           & \textbf{Inferior}             & \textbf{Equal}                            & \textbf{Outperform} \\
            \toprule
            \textbf{Score}              & \begin{math}< 95\%\end{math}  & \begin{math}\in [95\%, 105\%]\end{math}   & \begin{math}> 105\%\end{math} \\
            \textbf{Standard Deviation} & \begin{math}> 105\%\end{math} & \begin{math}\in [95\%, 105\%]\end{math}   & \begin{math}< 95\%\end{math} \\
            \bottomrule
        \end{tabular}
    }
    \label{tab:metrics}
\end{table}

In this way, we can easily compare the performance and stability of DSQN with that of the conversion-based SNN by the normalized scores and standard deviations.
Due to the generally poor stability of DRL algorithms, we developed Table \ref{tab:metrics} to comparing DSQN and the conversion-based SNN with the vanilla DQN based on the normalized scores and standard deviations.

\subsubsection{Code}
We implemented DSQN based on SpikingJelly \cite{SpikingJelly}, which is an open-source\footnote{The open-source code of SpikingJelly could be accessed at \url{https://github.com/fangwei123456/spikingjelly}.} deep learning framework for SNNs based on PyTorch.
The trained vanilla DQN was converted to SNN through the method proposed by \cite{Tan} based on their open-source code\footnote{The open-source code of \cite{Tan} could be accessed at \url{https://github.com/WeihaoTan/bindsnet-1}.}.

The source code could be accessed at our Github repository\footnote{\url{https://github.com/AptX395/Deep-Spiking-Q-Networks}}.

\subsection{Comparison results}

\begin{table*}[htbp]
    \centering
    \caption{The raw experimental results of the vanilla DQN, conversion-based SNN, and DSQN on 17 top-performing Atari games.}
    \begin{threeparttable}
        \scalebox{1.05}{
            \begin{tabular}{ccrlcrlcrl}
                \multirow{2}{*}{\textbf{Game}} & \multicolumn{3}{c}{\textbf{Vanilla DQN}} & \multicolumn{3}{c}{\textbf{Conversion-based SNN} \tnote{1} (\begin{math}t = \bm{500}\end{math}) \tnote{2}} & \multicolumn{3}{c}{\textbf{Deep Spiking Q-Network} \tnote{1} (\begin{math}t = \bm{64}\end{math}) \tnote{2}} \\
                
                ~ & Score & \multicolumn{2}{c}{\begin{math}\pm\end{math}std (\% Score)} & Score & \multicolumn{2}{c}{\begin{math}\pm\end{math}std (\% Score)} & Score & \multicolumn{2}{c}{\begin{math}\pm\end{math}std (\% Score)} \\
                
                \toprule
                
                Atlantis & \begin{math}493343.3\end{math} & \begin{math}21496.9\end{math} & (\begin{math}4.4\%\end{math}) & \begin{math}460700.0\end{math} & \begin{math}17771.0\end{math} & (\begin{math}3.9\%\end{math}) & \begin{math}\bm{487366.7}\end{math} & \begin{math}14400.6\end{math} & (\begin{math}\bm{3.0\%}\end{math}) \\
                
                BeamRider & \begin{math}7414.1\end{math} & \begin{math}1943.3\end{math} & (\begin{math}26.2\%\end{math}) & \begin{math}6041.2\end{math} & \begin{math}2423.2\end{math} & (\begin{math}40.1\%\end{math}) & \begin{math}\bm{7226.9}\end{math} & \begin{math}2348.7\end{math} & (\begin{math}\bm{32.5\%}\end{math}) \\
                
                Boxing & \begin{math}96.1\end{math} & \begin{math}3.1\end{math} & (\begin{math}3.2\%\end{math}) & \begin{math}91.3\end{math} & \begin{math}7.0\end{math} & (\begin{math}7.6\%\end{math}) & \begin{math}\bm{95.3}\end{math} & \begin{math}3.7\end{math} & (\begin{math}\bm{3.8\%}\end{math}) \\
                
                Breakout & \begin{math}425.4\end{math} & \begin{math}74.2\end{math} & (\begin{math}17.5\%\end{math}) & \begin{math}364.6\end{math} & \begin{math}108.0\end{math} & (\begin{math}29.6\%\end{math}) & \begin{math}\bm{386.5}\end{math} & \begin{math}61.1\end{math} & (\begin{math}\bm{15.8\%}\end{math}) \\
                
                Crazy Climber & \begin{math}120516.7\end{math} & \begin{math}16126.6\end{math} & (\begin{math}13.4\%\end{math}) & \begin{math}113133.3\end{math} & \begin{math}28441.9\end{math} & (\begin{math}25.1\%\end{math}) & \begin{math}\bm{123916.7}\end{math} & \begin{math}19142.0\end{math} & (\begin{math}\bm{15.4\%}\end{math}) \\
                
                Gopher & \begin{math}10552.7\end{math} & \begin{math}3935.1\end{math} & (\begin{math}37.3\%\end{math}) & \begin{math}\bm{10670.7}\end{math} & \begin{math}4189.9\end{math} & (\begin{math}\bm{39.3\%}\end{math}) & \begin{math}10107.3\end{math} & \begin{math}4303.2\end{math} & (\begin{math}42.6\%\end{math}) \\
                
                Jamesbond & \begin{math}906.7\end{math} & \begin{math}982.2\end{math} & (\begin{math}108.3\%\end{math}) & \begin{math}621.7\end{math} & \begin{math}147.0\end{math} & (\begin{math}\bm{23.6\%}\end{math}) & \begin{math}\bm{1156.7}\end{math} & \begin{math}2699.4\end{math} & (\begin{math}233.4\%\end{math}) \\
                
                Kangaroo & \begin{math}4140.0\end{math} & \begin{math}1605.5\end{math} & (\begin{math}38.8\%\end{math}) & \begin{math}4520.0\end{math} & \begin{math}1691.8\end{math} & (\begin{math}\bm{37.4\%}\end{math}) & \begin{math}\bm{8880.0}\end{math} & \begin{math}4041.3\end{math} & (\begin{math}45.5\%\end{math}) \\
                
                Krull & \begin{math}9309.3\end{math} & \begin{math}1072.5\end{math} & (\begin{math}11.5\%\end{math}) & \begin{math}7425.3\end{math} & \begin{math}2588.9\end{math} & (\begin{math}34.9\%\end{math}) & \begin{math}\bm{9940.0}\end{math} & \begin{math}999.5\end{math} & (\begin{math}\bm{10.1\%}\end{math}) \\
                
                Name This Game & \begin{math}11004.0\end{math} & \begin{math}1257.5\end{math} & (\begin{math}11.4\%\end{math}) & \begin{math}10541.0\end{math} & \begin{math}1653.9\end{math} & (\begin{math}15.7\%\end{math}) & \begin{math}\bm{10877.0}\end{math} & \begin{math}1581.2\end{math} & (\begin{math}\bm{14.5\%}\end{math}) \\
                
                Pong & \begin{math}20.2\end{math} & \begin{math}1.0\end{math} & (\begin{math}4.9\%\end{math}) & \begin{math}18.5\end{math} & \begin{math}1.4\end{math} & (\begin{math}7.3\%\end{math}) & \begin{math}\bm{20.3}\end{math} & \begin{math}0.9\end{math} & (\begin{math}\bm{4.6\%}\end{math}) \\
                
                Road Runner & \begin{math}54596.7\end{math} & \begin{math}6082.0\end{math} & (\begin{math}11.1\%\end{math}) & \begin{math}43160.0\end{math} & \begin{math}16322.1\end{math} & (\begin{math}37.8\%\end{math}) & \begin{math}\bm{48983.3}\end{math} & \begin{math}5903.1\end{math} & (\begin{math}\bm{12.1\%}\end{math}) \\
                
                Space Invaders & \begin{math}2274.5\end{math} & \begin{math}808.2\end{math} & (\begin{math}35.5\%\end{math}) & \begin{math}1387.3\end{math} & \begin{math}791.6\end{math} & (\begin{math}57.1\%\end{math}) & \begin{math}\bm{1832.2}\end{math} & \begin{math}735.4\end{math} & (\begin{math}\bm{40.1\%}\end{math}) \\
                
                Star Gunner & \begin{math}51070.0\end{math} & \begin{math}9513.2\end{math} & (\begin{math}18.6\%\end{math}) & \begin{math}1176.7\end{math} & \begin{math}2997.9\end{math} & (\begin{math}254.8\%\end{math}) & \begin{math}\bm{57686.7}\end{math} & \begin{math}6296.3\end{math} & (\begin{math}\bm{10.9\%}\end{math}) \\
                
                Tennis & \begin{math}-1.0\end{math} & \begin{math}0.0\end{math} & (\begin{math}0.0\%\end{math}) & \begin{math}-1.0\end{math} & \begin{math}0.0\end{math} & (\begin{math}0.0\%\end{math}) & \begin{math}-1.0\end{math} & \begin{math}0.0\end{math} & (\begin{math}0.0\%\end{math}) \\
                
                Tutankham & \begin{math}187.4\end{math} & \begin{math}60.3\end{math} & (\begin{math}32.2\%\end{math}) & \begin{math}190.9\end{math} & \begin{math}34.5\end{math} & (\begin{math}\bm{18.1\%}\end{math}) & \begin{math}\bm{194.7}\end{math} & \begin{math}51.4\end{math} & (\begin{math}26.4\%\end{math}) \\
                
                Video Pinball & \begin{math}316428.0\end{math} & \begin{math}223159.8\end{math} & (\begin{math}70.5\%\end{math}) & \begin{math}266940.1\end{math} & \begin{math}192004.0\end{math} & (\begin{math}71.9\%\end{math}) & \begin{math}\bm{275342.8}\end{math} & \begin{math}177157.0\end{math} & (\begin{math}\bm{64.3\%}\end{math}) \\
                
                \bottomrule
            \end{tabular}
        }
        \label{tab:rawResults}
        \begin{tablenotes}
            \footnotesize
            \item[1] The bold part represents the better among DSQN and the conversion-based SNN.
            Each game was run for 30 rounds.
            
            \item[2] The length of simulation time window \begin{math}t\end{math} of the conversion-based SNN and DSQN were set to \begin{math}\bm{500}\end{math} and \begin{math}\bm{64}\end{math} timesteps, respectively.
            The \textit{percentile} of the conversion-based SNN was set to \begin{math}\bm{99.9}\end{math} constantly.
        \end{tablenotes}
    \end{threeparttable}
\end{table*}

Table \ref{tab:rawResults} reports the raw experimental results of the three RL agents on 17 top-performing Atari games, including scores, standard deviations.
The sorting in Table \ref{tab:rawResults} is based on the lexicographical order of game names.
By using the metrics illustrated in the previous section, we obtained Table \ref{tab:normalizedResults} from Table \ref{tab:rawResults}, which shows the performance and stability difference between DSQN and the conversion-based SNN by normalizing the raw experimental results in Table \ref{tab:rawResults} with the scores and standard deviations of the vanilla DQN.
The sorting in Table \ref{tab:normalizedResults} is based on the score differences between DSQN and Conversion-based SNN.

\subsubsection{Performance}
According to the differences of the scores shown in Table \ref{tab:normalizedResults}, DSQN achieved higher scores than the conversion-based SNN on 15 out of 17 Atari games.
This illustrates the superiority of our direct learning method for DSQN compared to the conversion method.

At the same time, based on the scores shown in Table \ref{tab:normalizedResults} and the judging criteria in Table \ref{tab:metrics}, DSQN outperforms the vanilla DQN on 4 games, is equal to it on 9 game, and is inferior to it on 4 games.
This illustrates that our directly-trained Deep Spiking Reinforcement Learning architecture has the same level performance of the vanilla DQN in solving DRL problems.


\subsubsection{Stability}
According to the differences of the standard deviations shown in Table \ref{tab:normalizedResults}, DSQN achieved lower standard deviations than the conversion-based SNN on 12 out of 17 Atari games.
This illustrates that DSQN is stronger than the conversion-based SNN in terms of stability.

At the same time, based on the standard deviations shown in Table \ref{tab:normalizedResults} and the judging criteria in Table \ref{tab:metrics}, DSQN outperforms the vanilla DQN on 6 games, is equal to it on 2 games, and is inferior to it on 9 games.
This illustrates that DSQN has the same level stability of the vanilla DQN.

Furthermore, DSQN shows stronger generalization than the conversion-based SNN.
With the respective hyperparameters of DSQN and the conversion-based SNN being fixed, the experimental results show that DSQN outperforms the conversion-based SNN on most games.
This indicates that DSQN is more adaptable to different game environments, and its generalization is stronger than that of the conversion-based SNN.

\begin{table*}[htbp]
    \centering
    \caption{The performance and stability difference between DSQN and the conversion-based SNN.}
    \begin{threeparttable}
        \scalebox{1.05}{
            \begin{tabular}{ccccccc}
                \multirow{2}{*}{\textbf{Game}} & \multicolumn{2}{c}{\textbf{Conversion-based SNN} (\begin{math}t = \bm{500}\end{math})} & \multicolumn{2}{c}{\textbf{Deep Spiking Q-Network} (\begin{math}t = \bm{64}\end{math}) \tnote{1}} & \multicolumn{2}{c}{\textbf{DSQN - Conversion-based SNN} \tnote{2}} \\
                
                ~ & Score (\% DQN) & \begin{math}\pm\end{math}std (\% DQN) & Score (\% DQN) & \begin{math}\pm\end{math}std (\% DQN) & Score Difference & \begin{math}\pm\end{math}std Difference \\
                
                \toprule
                
                Star Gunner & \begin{math}2.3\%\end{math} & \begin{math}1367.7\%\end{math} & \begin{math}\bm{113.0\%}\end{math} & \begin{math}\bm{58.6\%}\end{math} & \begin{math}\bm{110.7\%}\end{math} & \begin{math}\bm{-1309.1\%}\end{math} \\
                
                Kangaroo & \begin{math}109.2\%\end{math} & \begin{math}96.5\%\end{math} & \begin{math}\bm{214.5\%}\end{math} & \begin{math}117.4\%\end{math} & \begin{math}\bm{105.3\%}\end{math} & \begin{math}20.9\%\end{math} \\
                
                Jamesbond & \begin{math}68.6\%\end{math} & \begin{math}21.8\%\end{math} & \begin{math}\bm{127.6\%}\end{math} & \begin{math}215.4\%\end{math} & \begin{math}\bm{59.0\%}\end{math} & \begin{math}193.6\%\end{math} \\
                
                Krull & \begin{math}79.8\%\end{math} & \begin{math}302.6\%\end{math} & \begin{math}\bm{106.8\%}\end{math} & \begin{math}\bm{87.3}\%\end{math} & \begin{math}\bm{27.0\%}\end{math} & \begin{math}\bm{-215.3\%}\end{math} \\
                
                Space Invaders & \begin{math}61.0\%\end{math} & \begin{math}160.6\%\end{math} & \begin{math}80.6\%\end{math} & \begin{math}113.0\%\end{math} & \begin{math}\bm{19.6\%}\end{math} & \begin{math}\bm{-47.6\%}\end{math} \\
                
                BeamRider & \begin{math}81.5\%\end{math} & \begin{math}153.0\%\end{math} & \begin{math}97.5\%\end{math} & \begin{math}124.0\%\end{math} & \begin{math}\bm{16.0\%}\end{math} & \begin{math}\bm{-29.0\%}\end{math} \\
                
                Road Runner & \begin{math}79.1\%\end{math} & \begin{math}339.5\%\end{math} & \begin{math}89.7\%\end{math} & \begin{math}108.2\%\end{math} & \begin{math}\bm{10.7\%}\end{math} & \begin{math}\bm{-231.3\%}\end{math} \\
                
                Pong & \begin{math}91.7\%\end{math} & \begin{math}151.3\%\end{math} & \begin{math}100.7\%\end{math} & \begin{math}95.6\%\end{math} & \begin{math}\bm{9.0\%}\end{math} & \begin{math}\bm{-55.7\%}\end{math} \\
                
                Crazy Climber & \begin{math}93.9\%\end{math} & \begin{math}187.9\%\end{math} & \begin{math}102.8\%\end{math} & \begin{math}115.4\%\end{math} & \begin{math}\bm{8.9\%}\end{math} & \begin{math}\bm{-72.5\%}\end{math} \\
                
                Atlantis & \begin{math}93.4\%\end{math} & \begin{math}88.5\%\end{math} & \begin{math}98.8\%\end{math} & \begin{math}\bm{67.8\%}\end{math} & \begin{math}\bm{5.4\%}\end{math} & \begin{math}\bm{-20.7\%}\end{math} \\
                
                Breakout & \begin{math}85.7\%\end{math} & \begin{math}169.8\%\end{math} & \begin{math}90.9\%\end{math} & \begin{math}\bm{90.6\%}\end{math} & \begin{math}\bm{5.1\%}\end{math} & \begin{math}\bm{-79.2\%}\end{math} \\
                
                Boxing & \begin{math}95.1\%\end{math} & \begin{math}238.4\%\end{math} & \begin{math}99.2\%\end{math} & \begin{math}120.0\%\end{math} & \begin{math}\bm{4.1\%}\end{math} & \begin{math}\bm{-118.4\%}\end{math} \\
                
                Name This Game & \begin{math}95.8\%\end{math} & \begin{math}137.3\%\end{math} & \begin{math}98.8\%\end{math} & \begin{math}127.2\%\end{math} & \begin{math}\bm{3.1\%}\end{math} & \begin{math}\bm{-10.1\%}\end{math} \\
                
                Video Pinball & \begin{math}84.4\%\end{math} & \begin{math}102.0\%\end{math} & \begin{math}87.0\%\end{math} & \begin{math}\bm{91.2\%}\end{math} & \begin{math}\bm{2.7\%}\end{math} & \begin{math}\bm{-10.8\%}\end{math} \\
                
                Tutankham & \begin{math}101.9\%\end{math} & \begin{math}56.1\%\end{math} & \begin{math}103.9\%\end{math} & \begin{math}\bm{82.0\%}\end{math} & \begin{math}\bm{2.0\%}\end{math} & \begin{math}25.9\%\end{math} \\
                
                Tennis & \begin{math}100.0\%\end{math} & \begin{math}100.0\%\end{math} & \begin{math}100.0\%\end{math} & \begin{math}100.0\%\end{math} & \begin{math}0.0\%\end{math} & \begin{math}0.0\%\end{math} \\
                
                Gopher & \begin{math}101.1\%\end{math} & \begin{math}105.3\%\end{math} & \begin{math}95.8\%\end{math} & \begin{math}114.2\%\end{math} & \begin{math}-5.3\%\end{math} & \begin{math}8.9\%\end{math} \\
                
                \midrule
                
                Average & \begin{math}83.8\%\end{math} & \begin{math}222.3\%\end{math} & \begin{math}106.3\%\end{math} & \begin{math}107.5\%\end{math} & \begin{math}22.5\%\end{math} & \begin{math}-114.8\%\end{math} \\
                
                \bottomrule
            \end{tabular}
        }
        \label{tab:normalizedResults}
        \begin{tablenotes}
            \footnotesize
            \item[1] The bold part in the two column of \textit{Deep Spiking Q-Network} represents that DSQN outperforms the vanilla DQN in terms of performance and stability.
            \item[2] The bold part in the two column of \textit{DSQN - Conversion-based SNN} represents that DSQN achieved higher scores and lower standard deviations than the conversion-based SNN.
        \end{tablenotes}
    \end{threeparttable}
\end{table*}

\begin{figure*}[htbp]
    \centering
    \subfloat[\begin{footnotesize}Scores of the vanilla DQN on Star Gunner.\end{footnotesize}]{
        \includegraphics[scale=0.35]{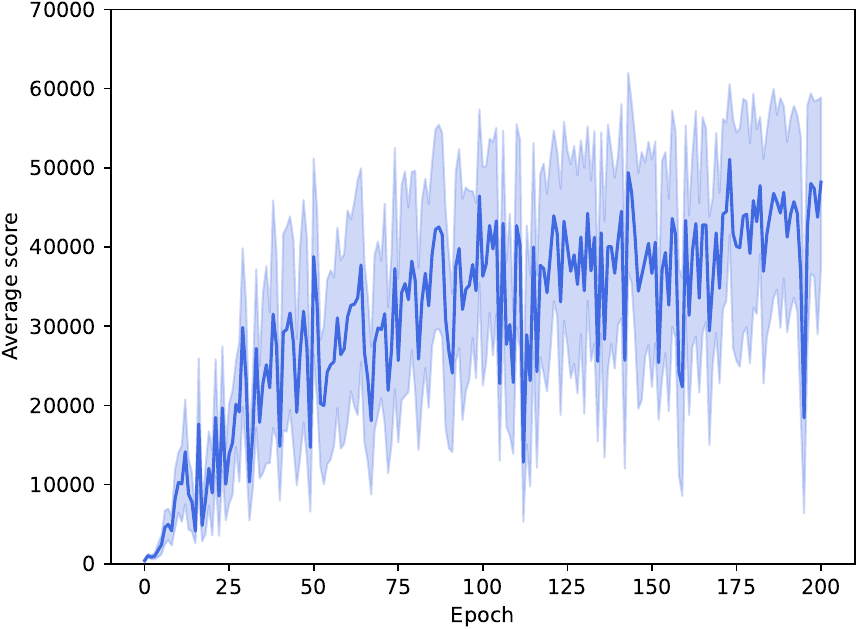}
        \label{subfig:dqnS}
    }
    \hfill
    \subfloat[\begin{footnotesize}Scores of DSQN on Star Gunner.\end{footnotesize}]{
        \includegraphics[scale=0.35]{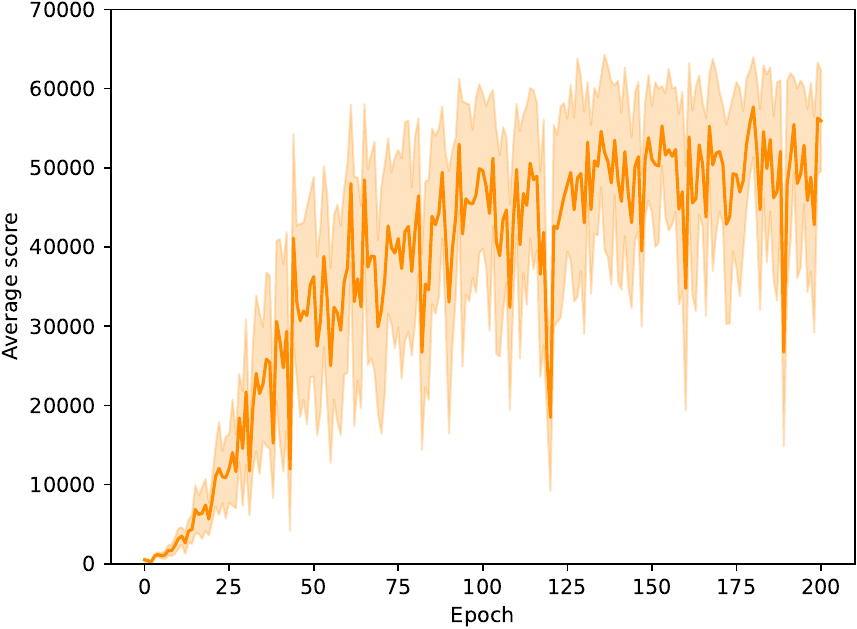}
        \label{subfig:dsqnS}
    }
    \hfill
    \subfloat[\begin{footnotesize}Average Q-values on Star Gunner.\end{footnotesize}]{
        \includegraphics[scale=0.35]{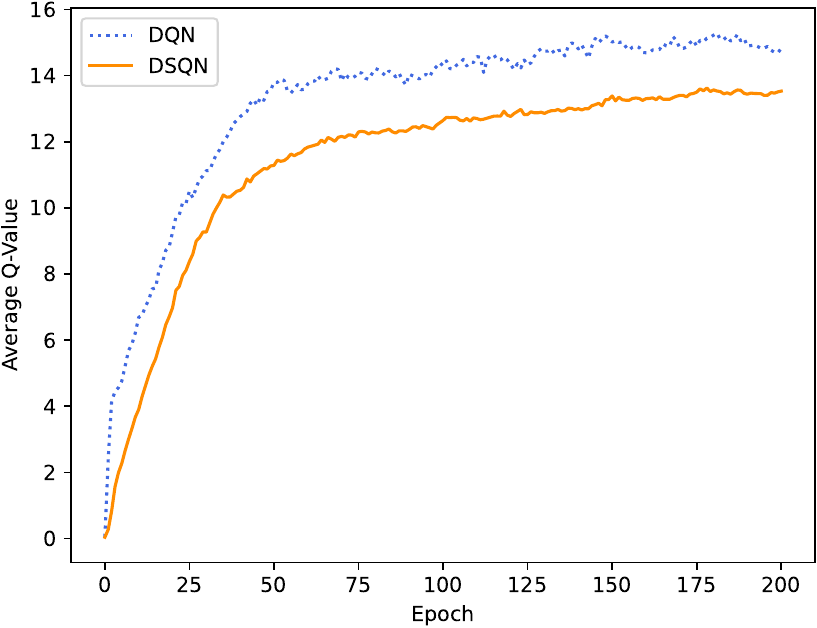}
        \label{subfig:starGunnerQ}
    }
    \\
    \subfloat[\begin{footnotesize}Scores of the vanilla DQN on Breakout.\end{footnotesize}]{
        \includegraphics[scale=0.35]{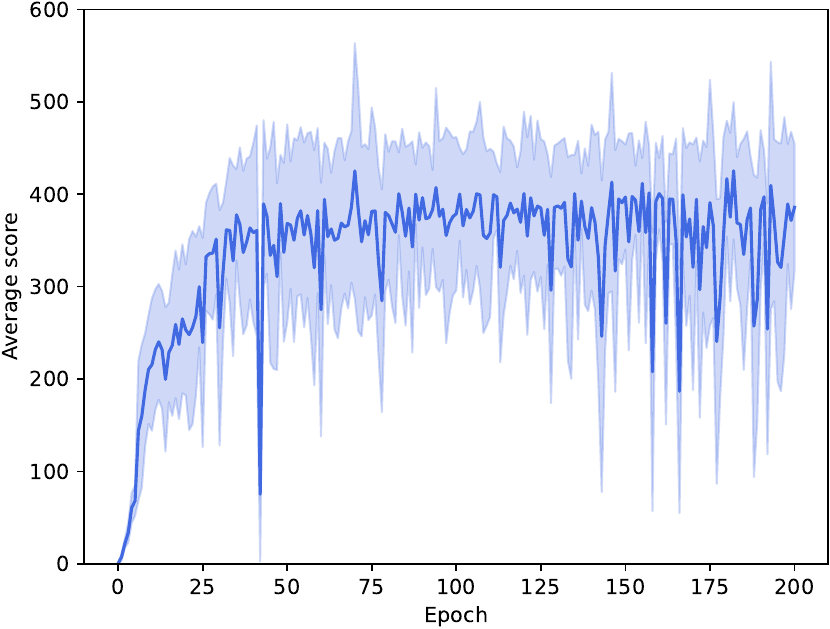}
        \label{subfig:dqnB}
    }
    \hfill
    \subfloat[\begin{footnotesize}Scores of DSQN on Breakout.\end{footnotesize}]{
        \includegraphics[scale=0.35]{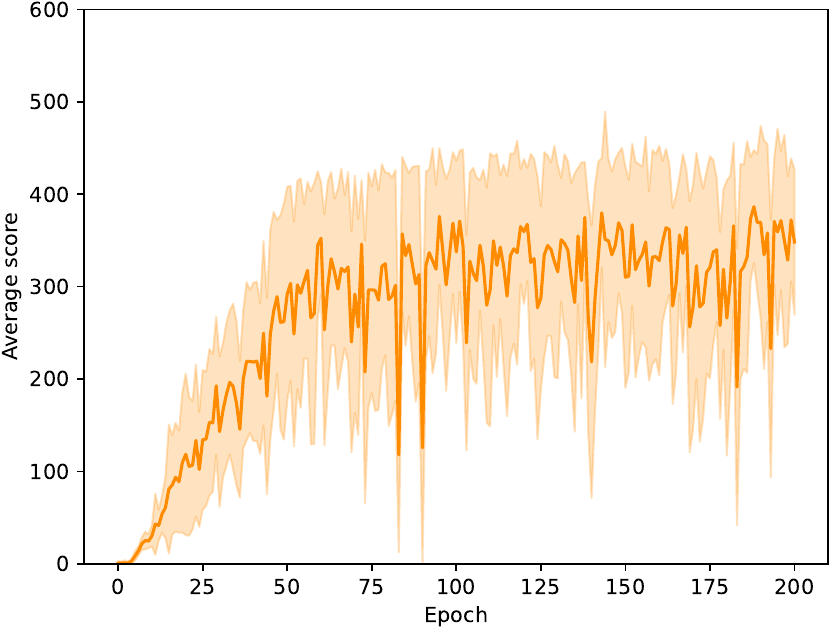}
        \label{subfig:dsqnB}
    }
    \hfill
    \subfloat[\begin{footnotesize}Average Q-values on Breakout.\end{footnotesize}]{
        \includegraphics[scale=0.35]{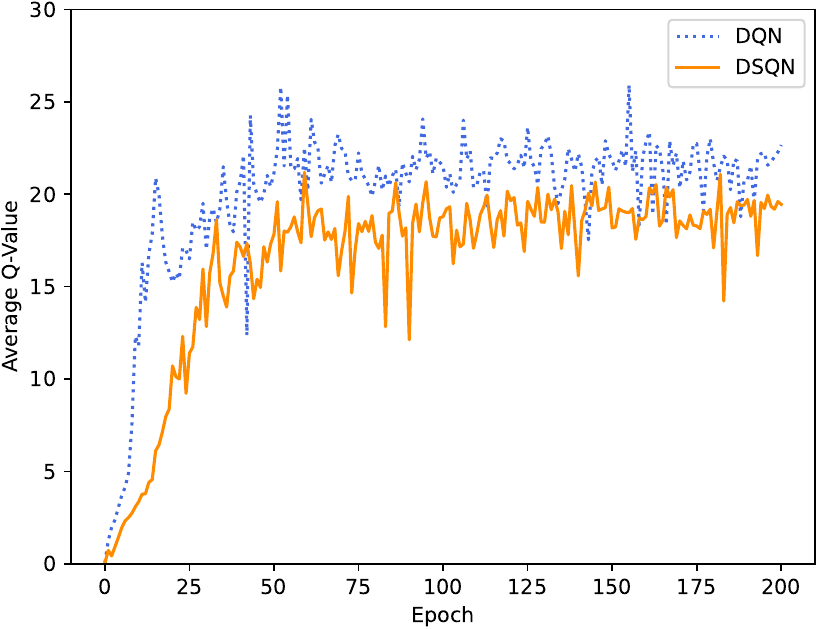}
        \label{subfig:breakoutQ}
    }
    \caption{The learning curves of DSQN and the vanilla DQN on Atari game Star Gunner and Breakout.
            During the training process, each episode was run for 30 rounds.
            In \ref{subfig:dqnS}, \ref{subfig:dsqnS}, \ref{subfig:dqnB}, and \ref{subfig:dsqnB}, each point is the average score achieved per episode.
            In \ref{subfig:starGunnerQ} and \ref{subfig:breakoutQ}, each point is the average max Q-value achieved per episode.
            (epoch = 250000 timesteps)}
    \label{fig:learningCurve}
\end{figure*}

\subsubsection{Learning capability}
In addition to the comparison of DSQN and the conversion-based SNN, we also analyzed the gap between DSQN and the vanilla DQN.

According to Table \ref{tab:normalizedResults}, the average score of DSQN slightly exceeds that of the vanilla DQN (i.e., 100\%), and the average standard deviation of DSQN also slightly exceeds that of the vanilla DQN.
This indicates that DSQN achieves the same level of learning capability as the vanilla DQN.

For instance, Figure \ref{fig:learningCurve} shows the learning curves of DSQN and the vanilla DQN on Atari game Star Gunner and Breakout during the training process.
As shown in Figure \ref{subfig:dqnS} and \ref{subfig:dsqnS}, the scores of DSQN are higher overall than that of the vanilla DQN.
As shown in Figure \ref{subfig:dqnB} and \ref{subfig:dsqnB}, although the scores of DSQN are slightly lower overall than that of the vanilla DQN, the standard deviations of DSQN are lower overall than that of the vanilla DQN.
These examples can confirm the learning capability of DSQN.
The curves of average Q-values shown in Figure \ref{subfig:starGunnerQ} and \ref{subfig:breakoutQ} can also confirm this.

\subsubsection{Energy-efficiency}
Besides, DSQN shows higher energy-efficiency than the conversion-based SNN as the fact that, DSQN achieved better performance than the conversion-based SNN while its simulation time window length is only \begin{math}\bm{64}\end{math} timesteps, which is one order of magnitude lower than that of the conversion-based SNN (\begin{math}\bm{500}\end{math} timesteps).

To compare the energy-efficiency of DSQN and the conversion-based SNN more precisely, we calculate their computational costs in the inference stage based on the number of synaptic operations using the method in \cite{Deng}, and we count the spiking times of them each time a decision is made on average.

For DSQN, let \begin{math}N _{D}\end{math} denotes the number of neurons in the neural network.
For each neuron, there are three addition and one multiplication operations, for a total of four synaptic operations according to Equation \eqref{eq:LIFu} and \eqref{eq:LIFv}.
Since the simulation time window length is \begin{math}64\end{math} timesteps, the computational cost is \begin{math}N _{D} \times 4 \times 64 = 256 N _{D}\end{math}.

For the conversion-based SNN, let \begin{math}N _{C}\end{math} denotes the number of neurons in the neural network.
For each neuron, there are one addition, i.e., a total of one synaptic operation according to Equation \eqref{eq:IFv}.
Since the simulation time window length is \begin{math}500\end{math} timesteps, the computational cost is \begin{math}N _{C} \times 1 \times 500 = 500 N _{C}\end{math}.

On the other hand, from the perspective of energy transfer, we take the two games shown in Figure \ref{fig:learningCurve} (Breakout and Star Gunner) as examples, and count the average spiking times of DSQN and the conversion-based SNN each time a decision is made, which are shown in Table \ref{tab:spiking_times}. From the results, the average spiking times of DSQN are only \begin{math}85.9\%\end{math} and \begin{math}85.5\%\end{math} of that of the conversion-based SNN on the two games, respectively.

\begin{table}[htbp]
    \centering
    \caption{Average spiking times of DSQN and the conversion-based SNN.}
    \begin{tabular}{ccc}
        \textbf{Game}   & \textbf{Conversion-based SNN} & \textbf{Deep Spiking Q-Network} \\
        \toprule
        Breakout        & 185.8k                        & \textbf{159.6k} \\
        Star Gunner     & 183.6k                        & \textbf{157.0k} \\
        \bottomrule
    \end{tabular}
    \label{tab:spiking_times}
\end{table}

According to the above analysis, when the structure and scale of DSQN and the conversion-based SNN are the same (i.e., \begin{math}N _{D} = N _{C}\end{math}), the energy-efficiency of DSQN is higher than that of the conversion-based SNN.

\subsubsection{In summary}
Combining the experimental results of DSQN and the conversion-based SNN in terms of both the scores and standard deviations, DSQN not only achieves higher scores than the conversion-bsed SNN on the vast majority of Atari games, but also achieves lower standard deviations at the same time.
This indicates that DSQN outperforms the conversion-based SNN in both performance and stability.
In addition, according to the comparison of DSQN and the vanilla DQN, even though DSQN is a SNN, its learning capability is not inferior to the vanilla DQN.

These experimental results demonstrated the superiority of DSQN over the conversion-based SNN in terms of performance, stability, generalization, and energy-efficiency.
Meanwhile, DSQN reaches the same level as DQN in terms of performance and surpasses DQN in terms of stability.

The code of the 

\subsection{Further Comparison}

To further explore the performance of our method, we compare our method with the latest variants of DQN, the Double DQN \cite{Weng}, and the state-of-the-art DRL method CDQN \cite{Wang1}. For fair comparison, we adapt our proposed deep spiking reinforcement learning strategy to the comparison baselines, named Double DSQN and CDSQN.

The Double DQN and DSQN are trained on 1M timesteps, and the CDQN and CDSQN are trained on 10M timesteps. The hyperparameters of the Double DQN and CDQN follow their references \cite{Weng} and \cite{Wang1} respectively.  Other experimental setups are the same as previous experiments.

The code of baselines are in the open source github \footnote{\url{https://github.com/thu-ml/tianshou}}\textsuperscript{,}\footnote{\url{https://github.com/Z-T-WANG/ConvergentDQN}}. The experiments are conducted based on the OpenAI Gym\footnote{\url{https://github.com/openai/gym}}.



\begin{table}[htbp]
    \centering
    \caption{Comparison results of Double DQN and Double DSQN.}
    \scalebox{0.85}{
        \begin{tabular}{cccc}
            \textbf{Game} & \textbf{Double DQN} & \textbf{Double DSQN} & \textbf{Score(\% Double DQN)} \\
            \toprule
            Beam Rider & \begin{math}3617.3\end{math} & \begin{math}4611.1\end{math} & \begin{math}\bm{127.5\%}\end{math} \\
            Breakout & \begin{math}137.9\end{math} & \begin{math}360.9\end{math} & \begin{math}\bm{261.7\%}\end{math} \\
            Jamesbond & \begin{math}748.0\end{math} & \begin{math}691.5\end{math} & \begin{math}92.4\%\end{math} \\
            Kangaroo & \begin{math}4302.0\end{math} & \begin{math}8620.0\end{math} & \begin{math}\bm{200.4\%}\end{math} \\
            Krull & \begin{math}10825.1\end{math} & \begin{math}10054.1\end{math} & \begin{math}92.9\%\end{math} \\
            Road Runner & \begin{math}34275.0\end{math} & \begin{math}43925.0\end{math} & \begin{math}\bm{128.2\%}\end{math} \\
            Space Invaders & \begin{math}694.9\end{math} & \begin{math}1089.8\end{math} & \begin{math}\bm{156.8\%}\end{math} \\
            \midrule
            Average & - & - & \begin{math}\bm{151.4\%}\end{math} \\
            \bottomrule
        \end{tabular}
    }
    \label{tab:furtherComparison1}
\end{table}

\begin{table}[htbp]
    \centering
    \caption{Comparison results of CDQN and CDSQN.}
    \scalebox{1.0}{
        \begin{tabular}{cccc}
            \textbf{Game} & \textbf{CDQN} & \textbf{CDSQN} & \textbf{Score(\% CDQN)} \\
            \toprule
            Beam Rider & \begin{math}8084.0\end{math} & \begin{math}6913.5\end{math} & \begin{math}85.5\%\end{math} \\
            Breakout & \begin{math}337.8\end{math} & \begin{math}275.5\end{math} & \begin{math}81.6\%\end{math} \\
            Jamesbond & \begin{math}235.0\end{math} & \begin{math}491.7\end{math} & \begin{math}\bm{209.2\%}\end{math} \\
            Kangaroo & \begin{math}2360.0\end{math} & \begin{math}7000.0\end{math} & \begin{math}\bm{296.6\%}\end{math} \\
            Krull & \begin{math}8404.3\end{math} & \begin{math}8938.6\end{math} & \begin{math}\bm{106.4\%}\end{math} \\
            Road Runner & \begin{math}28200.0\end{math} & \begin{math}30537.5\end{math} & \begin{math}\bm{108.3\%}\end{math} \\
            Space Invaders & \begin{math}1178.1\end{math} & \begin{math}1238.3\end{math} & \begin{math}\bm{105.1\%}\end{math} \\
            \midrule
            Average & - & - & \begin{math}141.8\%\end{math} \\
            \bottomrule
        \end{tabular}
    }
    \label{tab:furtherComparison2}
\end{table}


The comparison results of the Double DQN and CDQN are shown in Table \ref{tab:furtherComparison1} and \ref{tab:furtherComparison2} respectively. From which we see that our method achieves the averaged percent score of 151.4\% in the Double DQN, and achieves 141.8\% in the CDQN comparison. This further demonstrates the effectiveness of our method. Specifically, in the Kangaroo game, our method gets scores of 8620 and 7000 in the Double DSQN and CDSQN respectively, which significantly outperforms conventional scores of 4302 and 2360. The conclusions are consistent with previous experiments.




\begin{figure}[htbp]
    \centering
    \subfloat[\begin{footnotesize}Jamesbond\end{footnotesize}]{
        \includegraphics[scale=0.275]{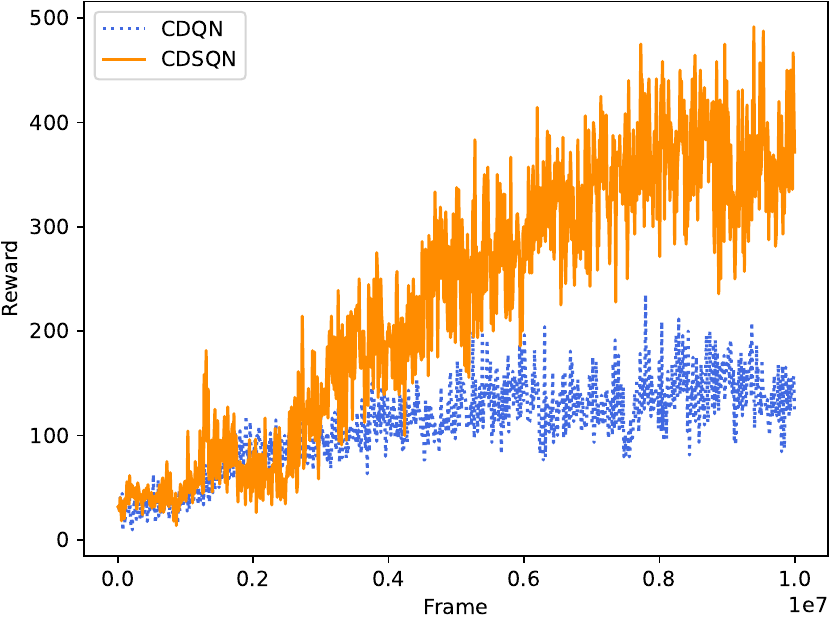}
        \label{subfig:supExJamesbond}
    }
    \hfill
    \subfloat[\begin{footnotesize}Kangaroo\end{footnotesize}]{
        \includegraphics[scale=0.275]{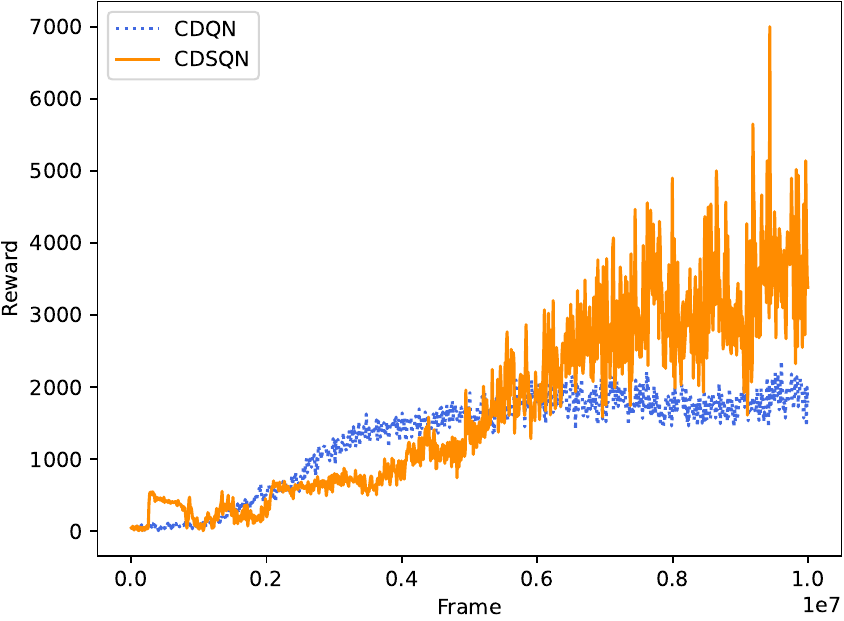}
        \label{subfig:supExKangaroo}
    }
    \\
    \subfloat[\begin{footnotesize}Krull\end{footnotesize}]{
        \includegraphics[scale=0.275]{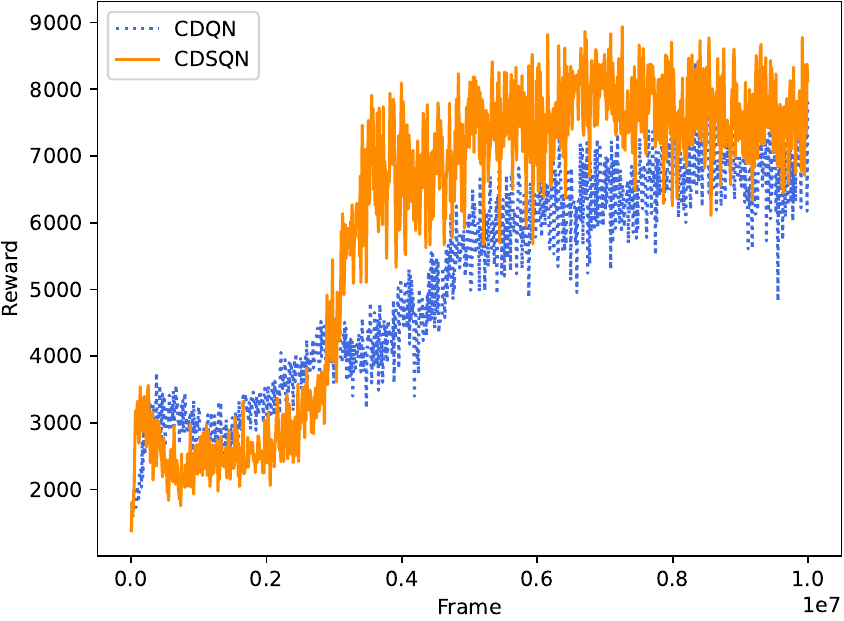}
        \label{subfig:supExKrull}
    }
    \hfill
    \subfloat[\begin{footnotesize}Road Runner\end{footnotesize}]{
        \includegraphics[scale=0.275]{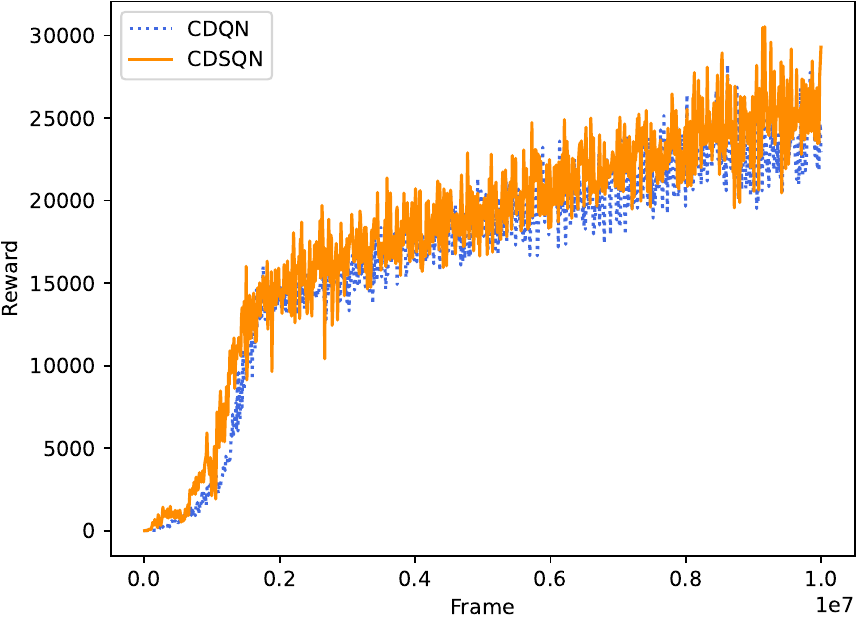}
        \label{subfig:supExRoadRunner}
    }
    \caption{The learning curves of CDQN and CDSQN on Atari game Jamesbond, Kangaroo, Krull, and Road Runner.}
    \label{fig:supEx}
\end{figure}

For further analysis, we plot the converge learning curves of the CDSQN and CDQN in the following Fig. 8, which demonstrate that our method outperforms conventional CDQN on most training frames.

\section{Conclusions and Future Works}
In this paper, we proposed a directly-trained Deep Spiking Reinforcement Learning architecture called Deep Spiking Q-Network (DSQN) to address the issue of solving Deep Reinforcement Learning (DRL) problems with SNNs.
To the best of our knowledge, our method is the first one to achieve state-of-the-art performance on multiple Atari games with directly-trained SNNs.
Our work serves as a benchmark for the directly-trained SNNs playing Atari games, and paves the way for future research to solving DRL problems with SNNs.
This work is designed for spiking deep Q-Learning based method. For other Deep Reinforcement Learning algorithms, such as Distributional RL, our work needs further study.
The theoretical analysis of LIF neuron's nature is not particularly in-depth, we could continue to improve it in the future work.
Moreover, based on the present work, and in order to better stimulate the potential of SNNs, we plan to conduct further research on actual neuromorphic hardware in the future.




\bibliographystyle{IEEEtran}
\bibliography{DSQN_Journal_Paper}
%


 




\begin{IEEEbiography}[{\includegraphics[scale=0.022]{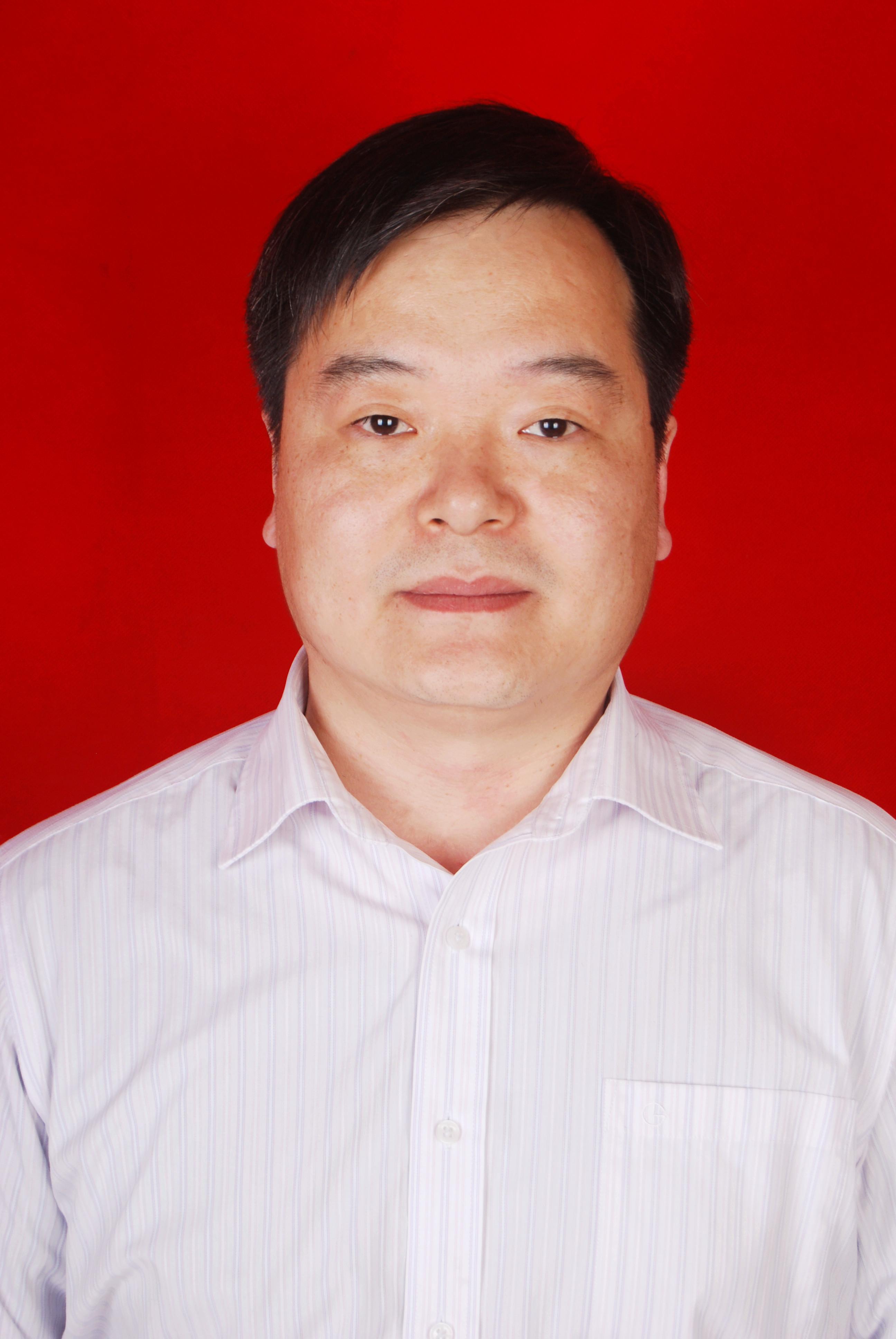}}]{Guisong Liu}
    received his B.S. degree in Mechanics from the Xi’an Jiao Tong University, Xi’an, China, in 1995, and his M.S. degree in Automatics, Ph.D degree in Computer Science both from the University of Electronic Science and Technology of China, Chengdu, China, in 2000 and 2007, respectively.
    Dr. Liu has filed over 20 patents, and published over 70 scientific conference and journal papers, and he was a visiting scholar at Humbolt University at Berlin in 2015.
    Before 2021, he is a professor in the School of Computer Science and Engineering, the University of Electronic Science and Technology of China, Chengdu, China.
    Now, he is a professor and the dean of the Economic Information Engineering School, Southwestern University of Finance and Economics, China.
    His research interests include pattern recognition, neural networks, and machine learning.
\end{IEEEbiography}

\begin{IEEEbiography}[{\includegraphics[scale=0.3]{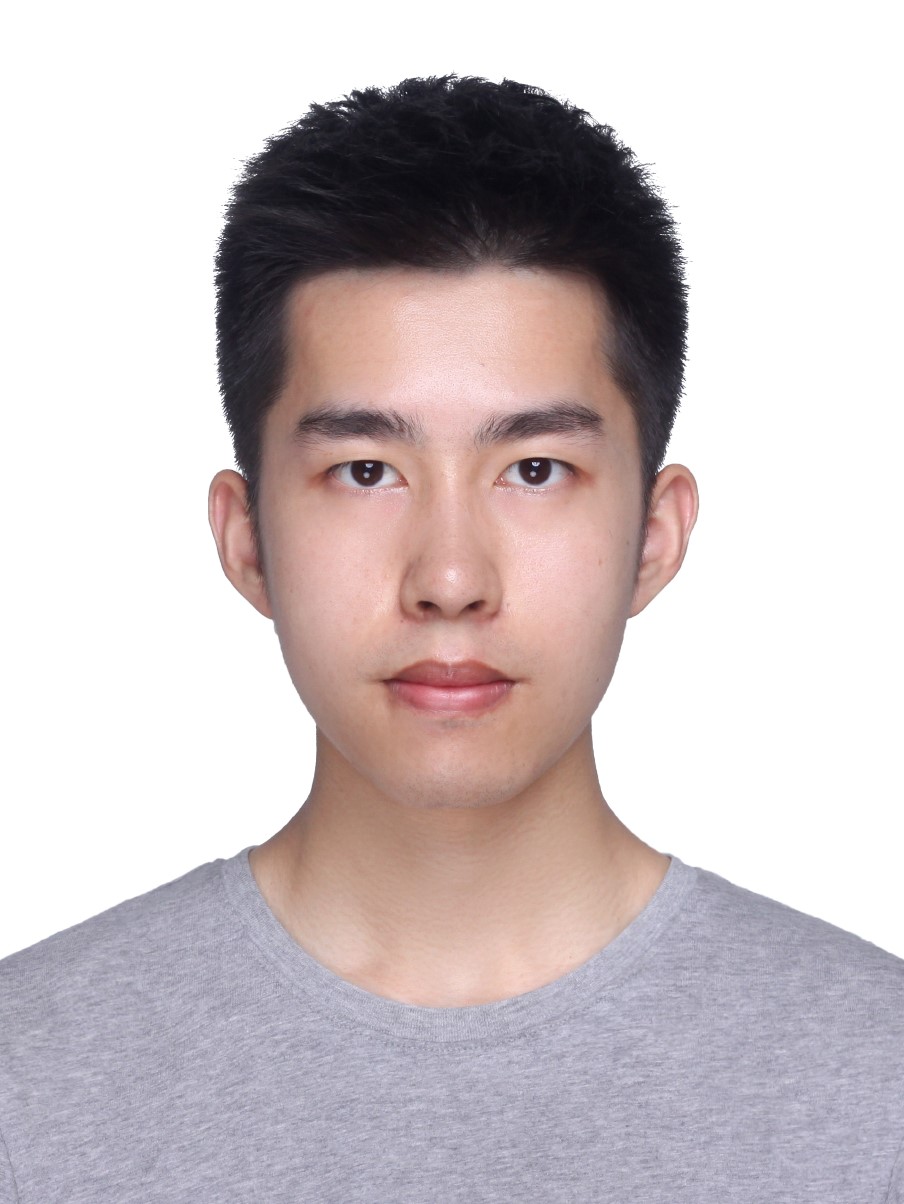}}]{Wenjie Deng}
    received his B.S. degree in Computer Science and Technology from Southwest University of Science and Technology, Mianyang, China, in 2020.
    He is pursuing the M.S degree in the School of Computer Science and Engineering, University of Electronic Science and Technology of China, Chengdu, China.
    His research interests include Neural Networks and Reinforcement Learning.
\end{IEEEbiography}

\begin{IEEEbiography}[{\includegraphics[scale=0.06]{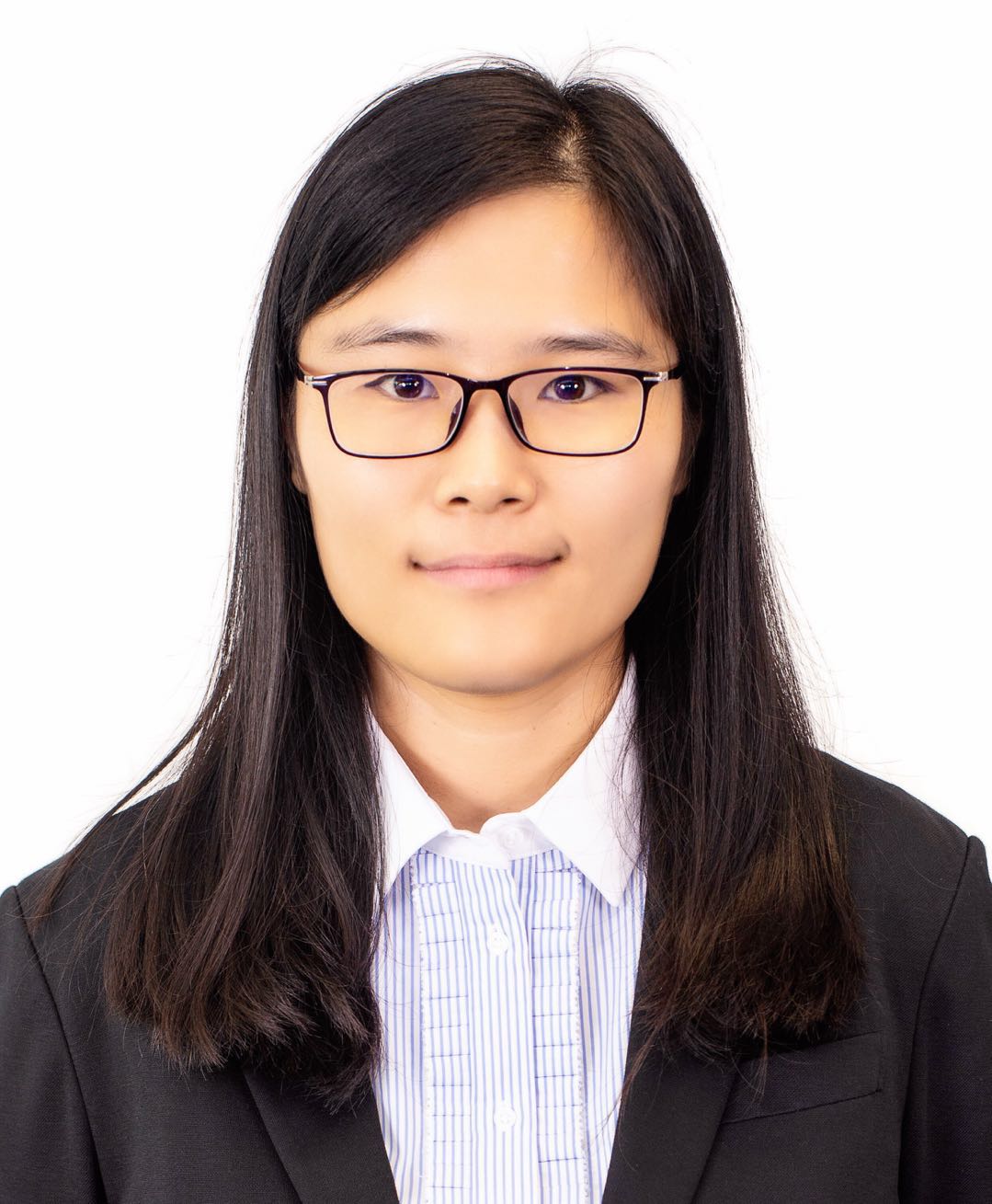}}]{Xiurui Xie}
    received the Ph.D. degree in Computer Science from the University of Electronic Science and Technology of China, Chengdu, China, in 2016.
    Dr. Xie worked as a Research Fellow in Nanyang Technological University, Singapore from 2017 to 2018, and worked as a Research Scientist in the Agency for Science, Technology and Research (ASTAR), Singapore from 2018 to 2020.
    Now, she is a associate professor in the University of Electronic Science and Technology of China.
    She has authored over 10 technical papers in prominent journals and conferences.
    Her primary research interests are neural networks, neuromorphic chips, transfer learning and pattern recognition.
\end{IEEEbiography}

\begin{IEEEbiography}[{\includegraphics[scale=0.05]{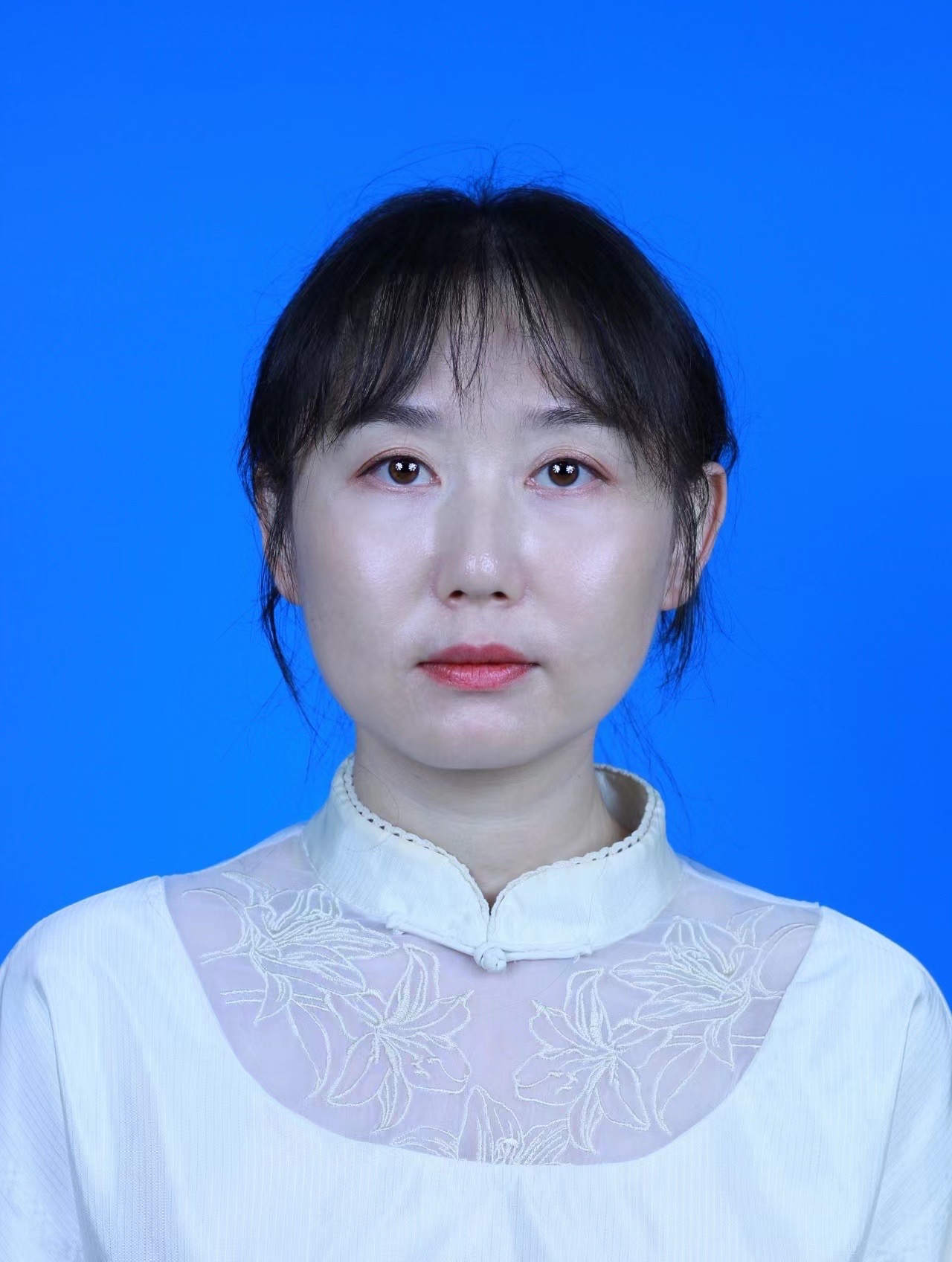}}]{Li Huang}
    received the Ph.D. degree in the School of Computer Science and Engineering from University of Electronic Science and Technology of China, supervised by Prof. Wenyu Chen.
    She is a lecturer at the School of Computing and Artificial Intelligence, Southwestern University of Finance and Economics, Chengdu, China.
    Her research interests include aspect sentiment analysis, machine translation, text summarization, and Continual Learning in Natural Language Processing.
\end{IEEEbiography}

\begin{IEEEbiography}[{\includegraphics[scale=0.35]{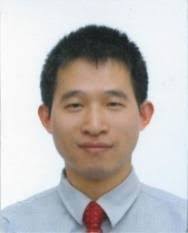}}]{Huajin Tang}
    received the B.Eng. degree from Zhejiang University, China in 1998, received the M.Eng. degree from Shanghai Jiao Tong University, China in 2001, and received the Ph.D. degree from the National University of Singapore, in 2005.
    He was a system engineer with STMicroelectronics, Singapore, from 2004 to 2006.
    From 2006 to 2008, he was a Post-Doctoral Fellow with the Queensland Brain Institute, University of Queensland, Australia.
    Since 2008, he was Head of the Robotic Cognition Lab, Institute for Infocomm Research, A*STAR, Singapore.
    Since 2014 he is a Professor with College of Computer Science, Sichuan University and now he is a Professor with College of Computer Science and Technology, Zhejiang University, China.
    He received the 2016 IEEE Outstanding TNNLS Paper Award and 2019 IEEE Computational Intelligence Magazine Outstanding Paper Award.
    His current research interests include neuromorphic computing, neuromorphic hardware and cognitive systems, robotic cognition, etc.
    Dr. Tang has served as an Associate Editor of IEEE Trans. on Neural Networks and Learning Systems, IEEE Trans. on Cognitive	and Developmental Systems, Frontiers in Neuromorphic Engineering, and Neural Networks.
    He was the Program Chair of IEEE CIS-RAM (2015, 2017), and ISNN (2019), and Co-Chair of IEEE Symposium on Neuromorphic Cognitive Computing (2016-2019).
    He is a Board of Governor member of International Neural Networks Society.
\end{IEEEbiography}


\end{document}